\documentclass[journal]{IEEEtran}
\usepackage{array}
\usepackage{graphicx} % picture
\usepackage{color} 
\usepackage{xcolor}
\usepackage{cite}
\usepackage{booktabs}
\usepackage{multirow}
\usepackage{amsmath}
\usepackage{amsfonts} %数学符号
\usepackage{amssymb} %数学符号
\usepackage{mathrsfs}%用于产生一种数学用的花体字
\usepackage{graphicx}
\usepackage{textcomp}
\usepackage{subfigure} %这个是自己添加的包
\usepackage{bm}%专门处理数学粗体的bm宏包

\usepackage{amsthm}

\usepackage{multirow}
\usepackage{subfigure}  %插入多图时用子图显示的宏包
\usepackage{color}
\usepackage{makecell}

\usepackage[ruled,linesnumbered]{algorithm2e}
\usepackage{algpseudocode}
\makeatletter
\newcommand{\removelatexerror}{\let\@latex@error\@gobble}
\makeatother

\begin{document}
\title{Incorporating Reachability Knowledge into a Multi-Spatial Graph Convolution Based Seq2Seq Model for Traffic Forecasting}
% Incorporating Reachability Knowledge into a Multi-Attention Spatio-Temporal Graph for Multi-Step Traffic Forecasting
\author{Jiexia~Ye, ~Furong~Zheng, ~Juanjuan~Zhao*,~Kejiang~Ye, IEEE~Member, ~Chengzhong~Xu, IEEE~Fellow\\
\thanks{*Corresponding author: Juanjuan Zhao}
\thanks{
 Jiexia Ye, Furong Zheng, Juanjuan Zhao, Kejiang Ye are with Shenzhen Institutes of Advanced Technology, Chinese Academy of Sciences, China (E-mail: \{jx.ye, fr.zheng, jj.zhao, kj.ye\}@siat.ac.cn).

Chengzhong Xu is with State Key Lab of IOTSC, Department of Computer Science, University of Macau, Macau SAR, China (E-mail: czxu@um.edu.mo).
}
}

\maketitle
\begin{abstract}
Accurate traffic state prediction is the foundation of transportation control and guidance. It is very challenging due to the complex spatiotemporal dependencies in traffic data. Existing works cannot perform well for multi-step traffic prediction that involves long future time period. The spatiotemporal information dilution becomes serve when the time gap between input step and predicted step is large, especially when traffic data is not sufficient or noisy. To address this issue, we propose a multi-spatial graph convolution based Seq2Seq model. Our main novelties are three aspects: (1) We enrich the spatiotemporal information of model inputs by fusing multi-view features (time, location and traffic states) (2) We build multiple kinds of spatial correlations based on both prior knowledge and data-driven knowledge to improve model performance especially in insufficient or noisy data cases. (3) A spatiotemporal attention mechanism based on reachability knowledge is novelly designed to produce high-level features fed into decoder of Seq2Seq directly to ease information dilution. Our model is evaluated on two real world traffic datasets and achieves better performance than other competitors.

\end{abstract}
% Note that keywords are not normally used for peer review papers.

\section{Introduction}
\IEEEPARstart{T}{raffic} forecasting is a key component of advanced traffic management systems, aiming to predict the future traffic states (e.g. traffic flow \cite{xie2020urban}, traffic speed \cite{ma2015long}, traffic time \cite{wang2018will}) in the traffic network. Accurate prediction can contribute to control traffic flow, allocate traffic resources, and ease traffic congestion\cite{9310691}. 

Multi-step traffic forecasting is a typical spatial-temporal forecasting problem and its has two main challenges. The first challenge is the complex spatiotemporal dependencies. The future traffic state of a region is influenced by many factors, such as its historical observation, the correlation with other regions, external factors (e.g. holiday, special events) and so on \cite{DBLP:conf/aaai/GuoLFSW19}. The correlations among regions are quiet complex. Although prior knowledge such as distance or travel time between regions might help to capture the spatial correlation, there are still some hidden patterns that need to be detected by data-driven methods. In addition, compared with one-step prediction task, long-term spatiotemporal correlation between input and output steps in multi-step scenario is more complex for that it changes as region, step and time gap between steps change \cite{DBLP:conf/iclr/LiYS018}. The second challenge is the model performance deterioration with insufficient qualified traffic data. Although we can collect more traffic data due to the update of transportation infrastructures, in many cases, the collected data is low quality with noise and some key features are missing. The available qualified traffic data is still insufficient. To the best of our knowledge, the traffic data used in most previous studies are less than one year \cite{DBLP:journals/corr/abs-1906-00560, DBLP:conf/mdm/GeLLZ19, DBLP:journals/corr/abs-1903-07789}, even one or two months \cite{DBLP:conf/ijcai/FangZMXP19}.

Recently, various deep learning based methods have been successfully applied to traffic prediction due to their superior capacities to capture complex traffic patterns. Graph neural networks (GNNs) are popular in extracting spatial dependency in traffic network \cite{DBLP:conf/mdm/GeLLZ19, DBLP:journals/tits/YuG19, DBLP:conf/aaai/GuoLFSW19,DBLP:conf/aaai/Diao0ZLXH19}.  Recurrent neural networks (RNNs) \cite{DBLP:journals/corr/abs-1903-05631, DBLP:conf/aaai/ChenLTZWWZ19, DBLP:conf/kdd/LiHCSWZP19} and Temporal convolution networks (TCNs) \cite{DBLP:conf/ijcai/WuPLJZ19, DBLP:conf/mdm/GeLLZ19, DBLP:conf/ijcai/YuYZ18} are often adopted to capture the temporal dependency. Seq2Seq model is widely utilized in multi-step forecasting \cite{DBLP:conf/ijcnn/ZhangWCC19, DBLP:journals/corr/abs-1911-08415, zhang2019multistep}. However, these previous works have the following limitations.

First, most existing works didn't capture the complex spatial correlations sufficiently. Many researchers pre-defined a matrix to reflect spatial correlation based on prior knowledge such as geometrical proximity, function similarity, transportation connectivity \cite{geng2019spatiotemporal, DBLP:conf/aaai/GuoLFSW19, guooptimized20}. However, any prior knowledge is limited and is not able to reflect some hidden traffic correlations among traffic regions. Some works designed an adaptive matrix to dig out spatial correlation on a data-driven basis \cite{DBLP:conf/ijcai/WuPLJZ19, DBLP:conf/icde/HuG0J19, DBLP:conf/aaai/Diao0ZLXH19}. However, when the data is not sufficient or noisy, the efficiency of data-driven method is deprecated and accurate prior knowledge might enhance the model performance in such situation. But most works only focus either pre-defined correlation or data-driven correlation for prediction.

Secondly, Seq2Seq model generally adopted in multi-step prediction has the information dilution problem \cite{DBLP:conf/iclr/LiYS018,DBLP:conf/ijcnn/ZhangWCC19,DBLP:journals/corr/abs-1911-08415}. When the original information on each input step arrives at a given output step, it has been diluted several times by both the encoder and decoder cell in Seq2Seq. Sufficient data might ease the dilution problem while insufficient data might amplify the dilution severity and decrease prediction performance. Therefore, modeling the long-term temporal correlation of traffic state between input and output steps to releviate dilution is particularly important.

In addition, each observed traffic feature (e.g. traffic speed, traffic flow) has both spatial attribute (i.e. its location) and temporal attributes (e.g. its time slot and week attribute). Most works \cite{DBLP:conf/uai/ZhangSXMKY18,DBLP:journals/corr/abs-1903-00919,DBLP:journals/access/ZhangYL19a, DBLP:conf/ijcai/FangZMXP19, 9207049, DBLP:journals/corr/abs-1903-07789} extracted the spatiotemporal patterns purely from traffic feature without making full use of its spatiotemporal attributes. However, such attributes can directly help model better identify spatiotemporal correlation among traffic states. They can also enrich the available spatiotemporal information when feature data is insufficient.

% most works purely fed the model with traffic feature (e.g. traffic speed \cite{DBLP:conf/uai/ZhangSXMKY18,DBLP:journals/corr/abs-1903-00919,DBLP:journals/access/ZhangYL19a}, traffic flow \cite{DBLP:conf/ijcai/FangZMXP19, 9207049, DBLP:journals/corr/abs-1903-07789}) without its spatial attribute (i.e. its location) and temporal attributes (e.g. its time slot and week attribute). However, when predicting traffic states in a network scale and at multiple future steps, such attributes can directly help model better identify spatiotemporal correlation among traffic states. They can also enrich the available information when data is insufficient. A few works utilized the temporal attributes by encoding them through one-hot embedding while it can't reflect the consecutive correlation of temporal attributes \cite{DBLP:conf/aaai/ZhangZQ17, DBLP:conf/mdm/GeLLZ19,DBLP:journals/corr/abs-1911-08415}.

To overcome the challenges and limitations above, this paper proposes a model MSGC-Seq2Seq, which mainly leverages the graph convolution and GRU based Seq2Seq to learn the complex spatiotemporal dependency from traffic data. Specifically, we fuse the traffic feature with its spatial attribute and temporal attribute to augment the spatiotemporal information of model input. To capture spatial correlation sufficiently, we first pre-define a matrix based on prior knowledge (i.e. geographical proximity and feature trend similarity) to extract the spatial correlation. Then, inspired by self-attention mechanism in transformer \cite{vaswani2017attention}, we calculate dynamic semantic attention between any two locations in the network on a data-driven basis.To tackle the dilution in Seq2Seq, we novelly design a attention mechanism based on reachability knowledge to model the cross-step spatial correlation. The high-level features produced are fed into the decoder at given output step directly without dilution by encoder. In addition, we integrate a multi-head temporal attention mechanism into Seq2Seq to allow the decoder to focus on the most relevant input steps related with the predicted output step, augmenting the useful input information for each output step prediction.

The main contributions of this paper are as follows:

\begin{itemize}
\item {We enrich the spatiotemporal information of model inputs by integrating the traffic state with its explicit spatiotemporal attributes via graph embedding methods.}

\item {We capture spatial correlations sufficiently from prior knowledge perspective (i.e. geographical proximity and feature trend similarity) and data-driven perspective (i.e. semantic similarity related with flow pattern).} 

\item {We novelly design a reachability based attention mechanism to construct output step related high-level features and feed them directly into the decoder to avoid the dilution by the encoder in Seq2Seq.} 

\item {Our MSGC-Seq2Seq model is evaluated on two real-world traffic datasets and performs better than other baselines, especially when data is insufficient and noisy. Our codes and datasets are public in https://github.com/start2020/MSGC-Seq2Seq.}

\end{itemize}

%%%%%%%%% Model
\begin{figure*}[htb]
%\centering
\includegraphics[width=0.98\textwidth, height=0.15\textheight]{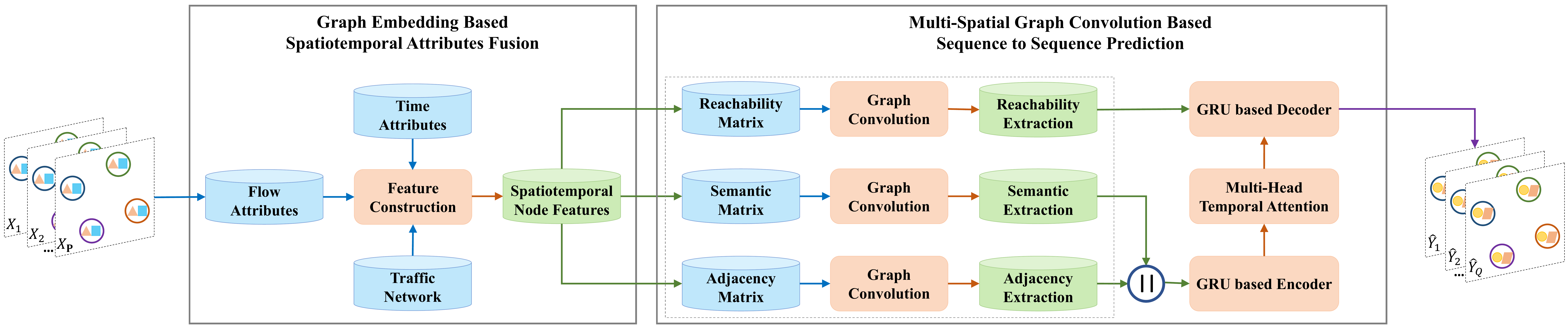}
\caption{The architecture of our model MSGC-Seq2Seq. The circles represent regions in the traffic network and the various shapes in the circles refer to various features. GRU refers to Gated Recurrent Unit.}
\label{fig:Model}
\end{figure*}

\section{Preliminaries}
\subsubsection{\textbf{Traffic Network as Graph}}
Traffic network is modeled as a graph $\mathbf{G} = (\mathbf{V},\mathbf{E},\mathbf{A})$ , which can be weighted or unweighted, directed or undirected. Each node in the graph represents a traffic region, such as a road sensor \cite{DBLP:conf/aaai/GuoLFSW19}, a road segment \cite{DBLP:conf/aaai/ChenLTZWWZ19}, a road intersection\cite{DBLP:journals/tits/YuG19}. $\mathbf{V}=\left\{v_{1}, \ldots, v_{\text{N}}\right\}$ refers $\text{N}$ traffic nodes on the traffic network. $\mathbf{E}$ is the edges set and each edge refers the connectivity between regions, which reflects some kind of spatial correlation, e.g. geographical proximity, semantic connectivity and reachability correlation.

\subsubsection{\textbf{Traffic Prediction Formulation}}
Assume that we can observe $\text{F}_{\text{I}}$ features of each node in the traffic graph during each given time period. The features of the whole network at time slot $p$ can be represented by a matrix $X_{p}=[x_{p}^{1},x_{p}^{2},\cdots,x_{p}^{\text{N}}] \in \mathbb{R}^{\text{N} \times \text{F}_{\text{I}}}$.
Further, the historical features of the whole network over the past $\text{P}$ time periods can be denoted as $[X_{1}, X_{2}, \cdots, X_{\text{P}}]\in \mathbb{R}^{\text{P}\times \text{N} \times \text{F}_{\text{I}}}$. In this paper, we aim to predict the future status of the whole network over $\text{Q}$ time period, denoted as $[Y_{1}, Y_{2}, \cdots, Y_{\text{Q}}]\in \mathbb{R}^{\text{Q} \times \text{N} \times \text{F}_\text{O}}$, where $Y_{q}=[y_{q}^{1},y_{q}^{2},\cdots, y_{q}^{\text{N}}] \in \mathbb{R}^{\text{N} \times \text{F}_\text{O}}$ and  $y^{j}_{q}\in \mathbb{R}^{\text{F}_\text{O}}$ represents the ground truth of region $j$ at future time period $q$. $\text{F}_\text{O}$ is the number of output features. We focus on multi-step prediction, therefore $\text{Q}>1$. 
This paper aims to find a function $f$ mapping the historical observations of the graph-based traffic network over past  $\text{P}$ time slices to its future observations over $\text{Q}$ time slices as follows:

\begin{small}
\begin{equation}
[\hat{Y}_{1},\cdots, \hat{Y}_{q}, \cdots, \hat{Y}_{\text{Q}}]= f([X_{1}, \cdots, X_{p}, \cdots, X_{\text{P}}])
\end{equation}
\end{small}where $\hat{Y}_{q}$ is the prediction output and its ground truth is $Y_{q}$.

%Figure \ref{fig:Model} depicts our model RSA-STGRU, which contains three blocks. The first block aims to capture the correlations among different regions across time. It calculates the attention scores that represent correlation between different nodes at different steps based on prior reachability information. The second block adopts idea from GNNs to aggregates the features of all nodes according to their importance (i.e. attention scores). Then it simply concatenates the aggregated features and utilizes the linear projection to reduce feature dimension to avoid overfitting. The reconstructed features have contained the dynamic spatial correlation and are fed into a GRU-based encoder-decoder framework to produce multi-step prediction for the target region. In the third block, there are two attention mechanisms integrated to improve prediction accuracy. The multi-feature attention can select the most relevant features for the whole prediction. The multi-temporal attention can distinguish the importance of different input steps for each future step.
%We have utilized valuable prior reachability knowledge to accurately filter ineffective features of some regions and retained the effective ones. Based on this along with inspiration from GNNs, we have reconstructed new features to containing dynamic spatial correlation and self-impact information at each input step,

 \begin{figure}[htbp]
\centering
\includegraphics[width=0.4\textwidth, height=0.12\textheight]{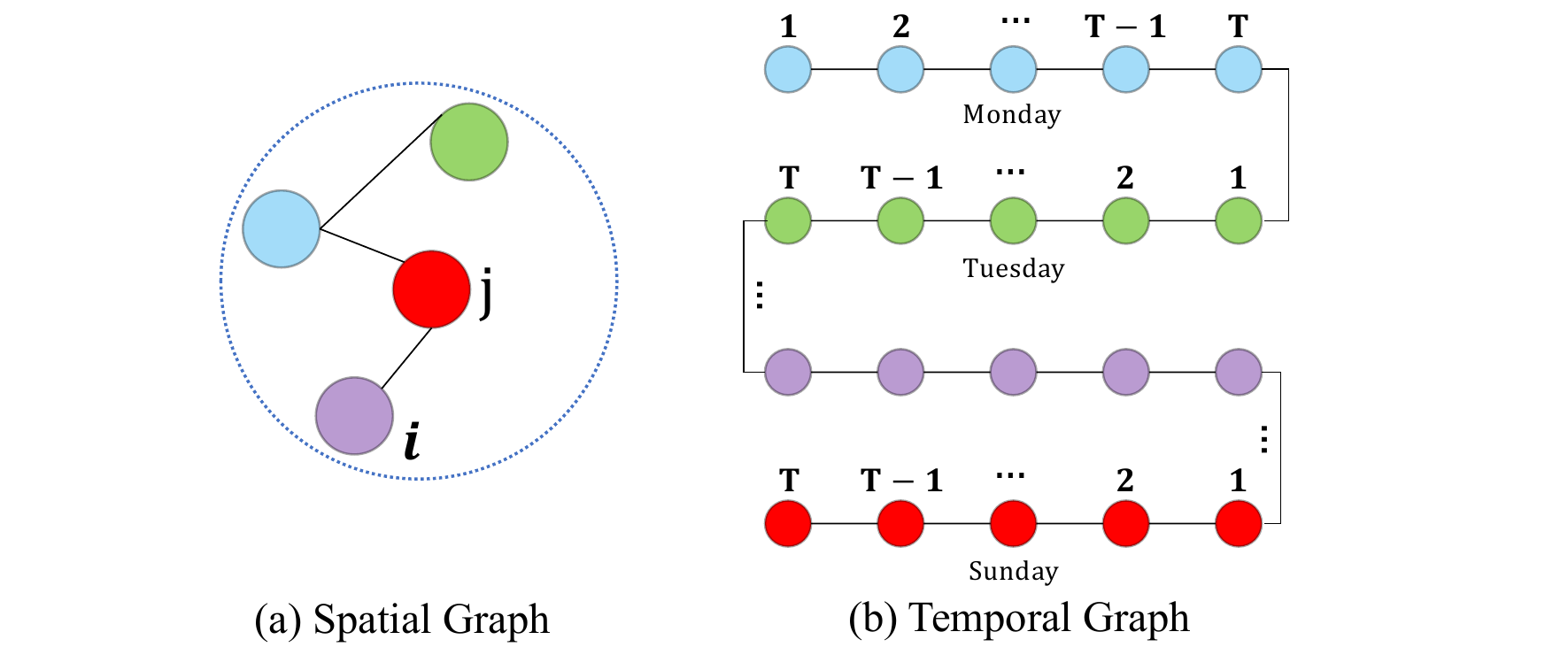}
\caption{Figure (a) is the spatial graph of the traffic network where a node represents a traffic region. Figure (b) is a temporal graph of a week where a node represents a time slot. Each day is divided equally into $\text{T}$ time slots, denoted as $[1, \dots, \text{T}]$. All time slots are connected sequentially as a line.}
\label{fig:LineGraph}
\end{figure}

\section{Model}
\subsection{\textbf{Overview}}
Our model MSGC-Seq2Seq (\underline{\textbf{M}}ulti-\underline{\textbf{S}}patial \underline{\textbf{G}}raph \underline{\textbf{C}}onvolution based \underline{\textbf{S}}equence to \underline{\textbf{S}}equence) (as shown in Figure \ref{fig:Model}) can be roughly divided into three stages. The first stage is to construct the model input with rich spatiotemporal information. We fuse the traffic feature with the spatiotemporal information extracted from its spatial attribute and temporal attribute via graph embedding methods. The second stage is to capture spatial correlations from multiple perspectives. We pre-define a spatial matrix based on geographical proximity and feature trend and we also develop a global spatial matrix on a data-driven basis. Both of them are both one-step correlation. In addition, a cross-step spatial correlation based on reachability knowledge is designed. Three graph convolution networks are leveraged to extract spatial correlations based on these spatial matrices. The third stage is to learn temporal dependency from the high-level features produced by graph convolution networks. High-level features based on one-step spatial correlations are merged and projected into the encoder of a Seq2Seq model to capture the temporal properties from these features. In addition, high-level features based on cross-step spatial correlation (i.e. reachability correlation) are fed into decoder directly to alleviate the information dilution problem. A temporal attention mechanism with multiple heads is integrated into Seq2Seq to enable the model focus on the most relevant input information. The details of our model is elaborated in the following sections.

\subsection{\textbf{Spatiotemporal Attributes Fusion}}
The traffic conditions in a traffic network have both spatial attribute and temporal attribute. The former refers to where the traffic data is observed (e.g. a sensor)  and the latter refers to when the traffic data is collected (e.g. time of day, day of week). Such spatiotemporal attributes of traffic data carry spatiotemporal information explicitly and are obviously valuable for traffic state prediction. In this paper, we want to enrich the model input by considering not only the traffic features but also their spatial and temporal attributes. We utilize the graph embedding methods to encode the spatial and temporal attributes and fuse them with traffic features.

%utilize the graph embedding methods to learn the local structure of each location and the sequential structure of each time slot. 

\textbf{Spatial Embedding.} Following \cite{DBLP:journals/corr/abs-1911-08415}'s work, we utilize node2vec \cite{DBLP:conf/kdd/GroverL16} to learn spatial embedding vector for each node in the traffic network based on its adjacent matrix as follows:
\begin{equation}
\text{SP} = \text{Node2vec}(\mathbf{A})
\end{equation}
where $\mathbf{A}$ is the adjacent matrix representing the topology of traffic network. $\text{SP} \in \mathbb{R}^{\text{N} \times \text{F}_{\text{S}}}$ and $\text{SP}_{i}\in \mathbb{R}^{ \text{F}_{\text{S}}}$ is the spatial embedding for traffic node $i$ and it has preserved the local structure information of this traffic node.

\textbf{Temporal Embedding.} Some works \cite{DBLP:journals/corr/abs-1911-08415} utilized one-hot embedding to learning temporal vector for each time slot. However,  one-hot embedding method can only distinguish time slots but overlooks the sequential correlation between time slots.  Different from previous works, we novelly flatten the time slots in a week and connect them sequentially as a line graph (as shown in Figure \ref{fig:LineGraph}). We utilize DeepWalk \cite{DBLP:conf/kdd/PerozziAS14} to learn embedding for each node (i.e. time slot) in the line graph. In this way, the temporal correlations among time slots are reserved in the embedding vector. The embedding method is as follows:
\begin{equation}
\text{TP} = \text{DeepWalk}(\mathbf{A}_{\text{T}})
\end{equation}
where $\mathbf{A}_{\text{T}}  \in \mathbb{R}^{\text{7T} \times \text{7T}}$ is the matrix of temporal line graph and $\text{T}$ is the number of total time slots in one day. $\text{7}$ refers to seven days in a week. $\text{TP} \in \mathbb{R}^{\text{7T} \times \text{F}_{\text{T}}}$ is the embedding matrix for all the time slots in a week.  $\text{TP}_{\text{dT+i}} \in \mathbb{R}^{\text{F}_{\text{T}}}$ is the temporal embedding for time slot $i$ in $d_{th}$ day of a week which carries the sequential temporal information.

\textbf{Features Fusion.} For each traffic node $i$ in time slot $t$, we can observe its traffic states $X_{t} ^{i} \in \mathbb{R}^{\text{F}_{\text{I}}}$.  The spatial attribute of traffic node $i$ is represented by $\text{SP}_{i}$ and the time slot $t$ has been assigned a temporal vector $\text{TP}_{t}$ according to its time of day and day of week attributes. We align them to the same dimension space through three fully connected layers respectively and then concatenate them together to get the fusion of all features as follows:
\begin{equation}
\text{XST}_{t}^{i} = \rho(W_{x}X_{t}) \| \rho(W_{t}\text{TP}_{t}) \| \rho(W_{i}\text{SP}_{i})
\end{equation}
where $\text{XST}_{t}^{i} \in \mathbb{R}^{3\text{F}_{\text{ST}}}$ is the fused features of traffic node $i$ at $t$. $\text{XST}_{t}=[\text{XST}_{t}^{1}, \dots, \text{XST}_{t}^{\text{N}}] \in \mathbb{R}^{\text{N} \times 3\text{F}_{\text{ST}}}$ is the fused matrix for the whole traffic network. $\rho$ is the activation function (e.g. ReLu). ${W_{x} \in \mathbb{R}^{\mathbf{F_{\text{ST}}} \times \text{F}_\text{I}},W_{t} \in \mathbb{R}^{\text{F}_{\text{ST}} \times \text{F}_{\text{T}}},W_{i} \in \mathbb{R}^{\text{F}_{\text{ST}} \times \text{F}_{\text{S}}} }$ are trainable parameters shared by all the traffic nodes. 

\subsection{\textbf{Multi-view Spatial Graph Convolutions}}
 \begin{figure}[htb]
\centering
\includegraphics[width=0.45\textwidth, height=0.12\textheight]{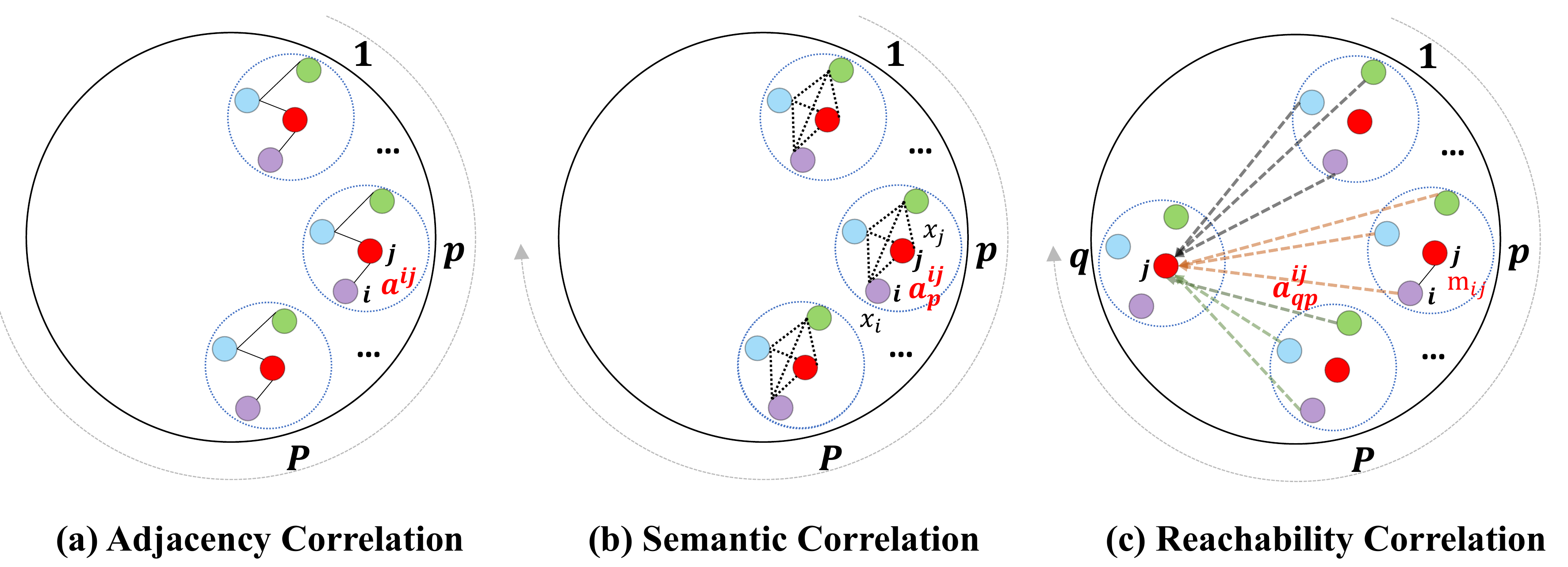}
\caption{The visualization of three spatial attention scores. $[1,\cdots, p, \cdots, P]$ are the input steps and $q$ is an output step. $a^{ij}$ describes the spatial correlation between traffic node $i$ and traffic node $j$ based on geographical proximity and feature trend similarity. $a^{ij}$ is static and local. $a^{ij}_{p}$ represents the semantic correlation of node pair $ij$ based on their flow patterns on time slot $p$. $a^{ij}_{p}$ is dynamic and global. $a^{ij}_{qp}$ extracts the reachability based spatial correlation determining by the travel time $m_{ij}$ of node pair and their time slot gap. $a^{ij}_{qp}$ is cross space and time.}
\label{fig:Dynamic}
\end{figure}
The fused features contain rich spatiotemporal information for each traffic node. However, the future traffic condition of a target traffic node is influenced not only by its own historical information but also by historical observations of other traffic nodes. The fact that traffic objects (e.g. passengers, vehicles) keep moving from one traffic node to another traffic node results in the spatial correlations between traffic nodes and such correlations are complex. Different from most previous works only capturing one kind of spatial correlation, we model the complex spatial dependency in the traffic network comprehensively from multiple perspectives, i.e. adjacency correlation, semantic correlation and reachability correlation. Following previous works \cite{geng2019spatiotemporal, DBLP:conf/mdm/GeLLZ19, DBLP:journals/tits/YuG19, DBLP:conf/aaai/GuoLFSW19,DBLP:conf/aaai/Diao0ZLXH19}, we leverage graph convolutional network to extract each spatial correlation respectively.

%Many previous works observe that directly connected regions are usually highly relevant to each other and they focus on extracting the dependency from adjacent neighbors which we refer as adjacent correlation. Some other works argue that those indirectly connected regions also affect the target regions for that they might have similar flow patterns and such dependency is referred as similarity correlation. All these works focus on capturing the spatial correlation between regions at the same time slot. However, we can observe one kind of spatial correlation related with both the regions and their time slots, i.e. reachability correlation. The reachability between any two regions depends on the travel time between them and the time slots which they are in. The reachability can measure the influence between regions from the traffic flow perspective, e.g. if one region is not reachable by another at a given time slot, they have little influence to each other. In this paper, we novelly extract the spatial correlation on traffic network based on reachability knowledge. Further,

\subsubsection{\textbf{Graph Convolution Network}}Graph Convolution generalizes convolution operation from regular grid data to graph data. It models influence from related nodes on the target node through aggregating the features of related nodes  \cite{DBLP:journals/corr/BrunaZSL13, DBLP:conf/nips/DefferrardBV16,DBLP:conf/iclr/KipfW17}. Such aggregation is usually achieved by the matrix multiplication of feature matrix and spatial matrix. The aggregated features are fed into fully connected layers to produce high-level features. The graph convolution layer adopted in this paper can be denoted as follows:
\begin{equation}
Y^{l}= \rho(W\text{L}X^{l-1})=\text{GCN}(A,X^{l-1})
\end{equation}
where $X^{l-1}$ is the input and $Y^{l}$ is the output of  $l_{th}$ graph convolution layer. $\rho$ is activation function and $W$ is trainable parameter. $\text{L}=I_{N}-D^{-\frac{1}{2}} A D^{-\frac{1}{2}}$ is the normalized version of spatial matrix $A$ which represents spatial correlation in the traffic network. Different spatial correlation is represented by different spatial matrix. Next, we introduce the construction of spatial matrix for each kind of spatial correlation in this paper.
%%%%%%%%%%%  figure:模型
\subsubsection{\textbf{Semantic Spatial Correlation}} 
% 语义就是点跟点之间的关系
If two traffic nodes share similar flow patterns, they might have similar traffic conditions, such as traffic congestion, functional similarity and owning similar POIs. Therefore, they can learn from each other. Such spatial correlation based on current flow patterns of traffic nodes is defined as semantic correlation.

\textbf{Dynamic Semantic Score.} Inspired by the transformer \cite{vaswani2017attention}, a state-of-the-art method to learn correlations between objects, we assign two latent semantic spaces to each traffic node as follows:
\begin{equation}
\begin{aligned}
\text{K}_{t}^{i}& = W_{k2}(\text{ReLu}(W_{k1}\text{XST}_{t}^{i})) \in \mathbb{R}^{F} \\
\text{Q}_{t}^{i}& = W_{q2}(\text{ReLu}(W_{q1}\text{XST}_{t}^{i})) \in \mathbb{R}^{F} \\
\end{aligned}
\end{equation}
where $\text{K}_{t}^{i}$, $\text{Q}_{t}^{i}$  are the embeddings in key space and query space respectively for traffic node $i$ at time slot $t$. $\{W_{k1},W_{k2},W_{q1},W_{q2}\}$ are trainable parameters shared by all traffic nodes. Further, we can calculate the influence from traffic node $j$ on traffic node $i$ at time slot $t$ as $a^{ij}_{t}= (\text{K}_{t}^{i})^{T}\boldsymbol{\cdot}\text{Q}_{t}^{j}\in \mathbb{R}$. The corresponding dynamic attention matrix is as follows:
\begin{equation}
\begin{split}
A^{f}_{t}=\text{Softmax}(
\left[\begin{array}{cccc}
{a^{11}_{t}} & {...} &{a^{1\text{N}}_{t}}\\
{...} & {...} &{...}\\
{a^{\text{N}1}_{t}} & {...} &{a^{\text{N}\text{N}}_{t}}\\
\end{array}\right])
\end{split}
\end{equation}
Note that the Softmax function is operated on each row.

\textbf{Semantic Correlation Extraction.} The dynamic attention matrix contains dynamic correlations between any two traffic nodes based on their flow patterns. We capture such spatial dependency through GCN at each input step $p$ as follows:
\begin{equation}
\text{XF}_{p} = \text{GCN}(A^{f}_{p}, \text{XST}_{p})
\end{equation}
where $\text{XF}_{p} \in \mathbb{R}^{\text{N} \times \text{F}_{\text{F}}}$ is the high-level output matrix of the traffic network.

\subsubsection{\textbf{Adjacent Trend Spatial Correlation}} Many previous works focus on capturing the influence between geographically adjacent traffic nodes and assign different weights to different neighbors of the target traffic node to model their different influence strength
 \cite{DBLP:conf/icde/HuG0J19, DBLP:conf/icpr/ZhangJCXP18, DBLP:conf/ijcnn/ZhangWCC19, DBLP:conf/uic/LiPLXDMWB18}. It is argued in  \cite{DBLP:conf/icde/HuLBCF16} that the correlation of two adjacent roads depends on their historical traffic speed trends. The larger the proportion that the speeds of two traffic nodes both rise or fall is, the higher the correlation between two adjacent traffic node is.

\textbf{Adjacent Trend Score.} Following \cite{DBLP:conf/icde/HuLBCF16}'s works, we define the adjacent trend score to present the weight between adjacent traffic nodes based on their historical traffic states trends as follows:
\begin{equation}
\begin{small}
a^{ij} = \frac{\sum_{f=1}^{\text{F}_{\text{I}}}[{\text{c}(v^{i}_{t,f} \geq \bar{v}^{i}_{f}, v^{j}_{t,f} \geq \bar{v}^{j}_{f})}+\text{c}(v^{i}_{t,f}<\bar{v}^{i}_{f}, v^{j}_{t,f}<\bar{v}^{j}_{f})]}{\text{F}_{\text{I}}*\textbf{Total}}
\end{small}
\end{equation}
where $v^{i}_{t,f}$ and $v^{j}_{t,f}$ are the $f_{th}$ traffic feature of traffic node $i$ and traffic node $j$ at time $t$. $\bar{v}^{i}_{f}$ and $\bar{v}^{j}_{f}$ are their average values respectively. $\text{c}(v^{i}_{t,f} \geq \bar{v}^{i}_{f}, v^{j}_{t,f} \geq \bar{v}^{j}_{f})$ and $\text{c}(v^{i}_{t,f}<\bar{v}^{i}_{f}, v^{j}_{t,f}<\bar{v}^{j}_{f})$ are the numbers of time slots that the $f_{th}$ traffic feature of adjacent traffic nodes both rise or fall. $\text{F}_{\text{I}}$ is the number of features and \textbf{Total} is the total number of time slots in historical data.  
For any two traffic nodes, if they are adjacent geographically, we calculate the adjacent trend score as their edge weight. Otherwise, their edge weight is set as zero. Based on the adjacent trend score, we obtain the adjacent trend matrix $\text{A}^{a}=(a_{ij})_{\text{N}\times \text{N}}$. Then we capture the spatial locality based on historical adjacent trend through GCN as follows:
\begin{equation}
\text{XA}_{p} = \text{GCN}(\text{A}^{a}, \text{XST}_{p})
\end{equation}
where $\text{XA}_{p} \in \mathbb{R}^{\text{N} \times \text{F}_{\text{A}}}$ contains local spatial dependency based on historical trends of traffic features.

 \begin{figure}[htb]
\centering
\includegraphics[width=0.45\textwidth, height=0.16\textheight]{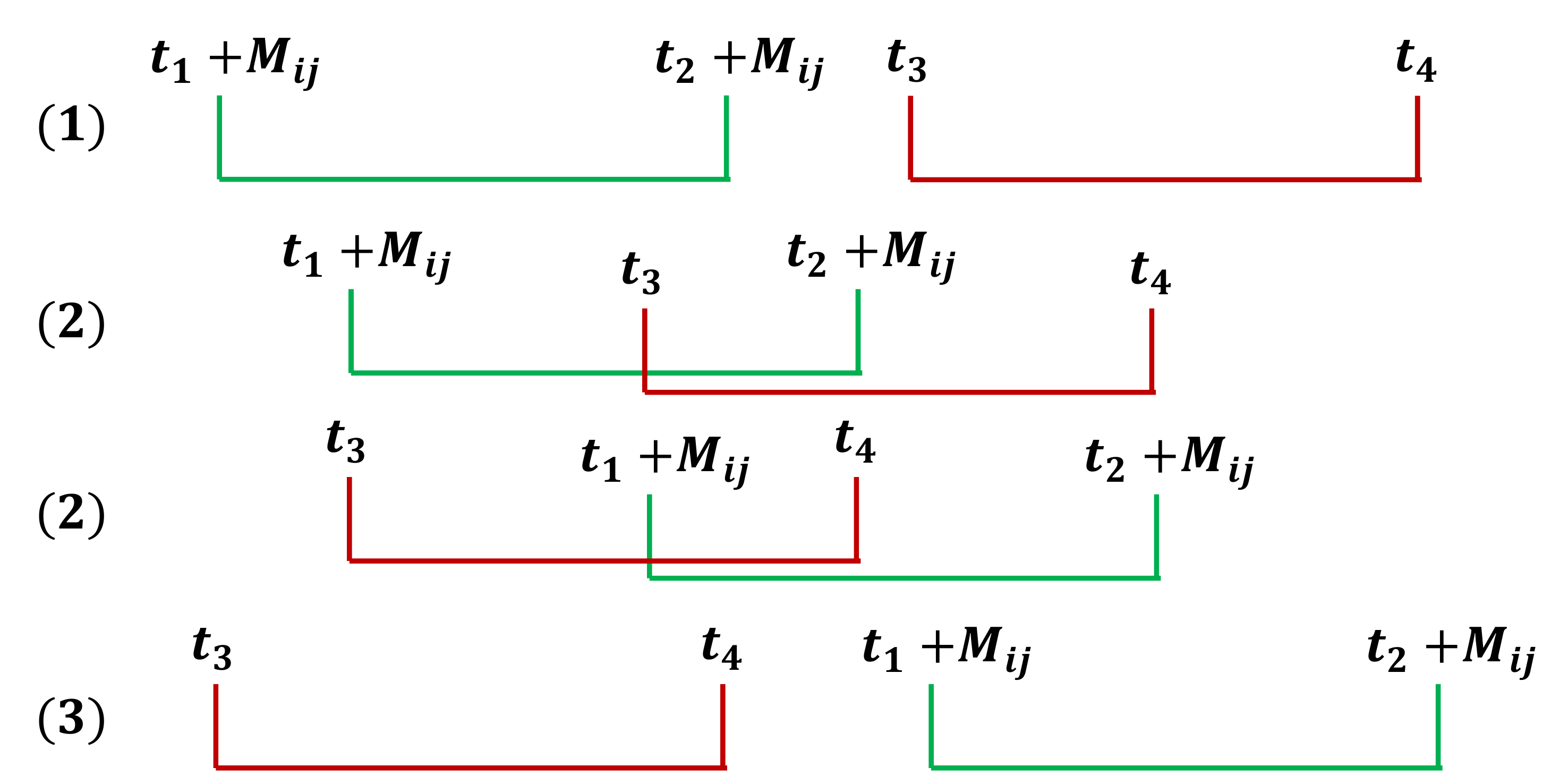}
\caption{For the passengers/vehicles departing from traffic node $i$ at time span $[t_1, t_2]$, most of them arrive at traffic node $j$ during time span $[t_1+\text{M}_\text{ij}, t_2+\text{M}_\text{ij}]$ and $\text{M}_\text{ij}$ is the average travel time between traffic node $i$ and traffic node $j$. The time span $[t_3, t_4]$ is the predicted time span. There are four correlations between predicted time span and arriving time span.}

\label{fig:TimeLine}
\end{figure}

\subsubsection{\textbf{Reachability Spatial Correlation Across Steps}} The reachability correlation intends to model the spatial correlation based on reachability knowledge between any traffic nodes from input steps to output steps.

\textbf{Reachability Knowledge.}
Suppose that we want to predict the traffic states $v_{j}$ of the target traffic node $j$ during interval $[t_3, t_4]$ with the traffic states $v_{i}$ of the context traffic node $i$ during previous interval $[t_1, t_2]$. We denote the average travel time from node $i$ to node $j$ as $\text{M}_\text{ij}$. For simplicity, we assume that all the passengers spend almost the same time from $i$ to $j$. For all passengers departing from context traffic node $i$ during $[t_1, t_2]$ to target traffic node $j$, we can observe that (as shown in Figure \ref{fig:TimeLine}):

(1) if $t_3 > t_2+\text{M}_\text{ij}$, nearly all passengers can't reach traffic node $j$ during $[t_3, t_4]$. Therefore, there is little correlation between traffic condition $v_{j}$ and $v_{i}$ from reachability perspective.

(2) if $t_4>t_2+\text{M}_\text{ij}\geq t_3$ or $t_4>t_1+\text{M}_\text{ij}\geq t_3$, we can infer that part of passengers can reach traffic node $j$ during $[t_3, t_4]$. For simplicity, we assume that all passengers enter a traffic node evenly during the given time slot, therefore the influence strength between these two traffic nodes depends on the overlap of $[t_1+\text{M}_\text{ij}, t_2+\text{M}_\text{ij}]$ and $[t_3, t_4]$.

 (3) if $t_1+\text{M}_\text{ij}>t_4$, passenger might have already arrived at traffic node $j$ before $t_4$. If passengers don't stay, they might leave $j$ for a long time. Therefore, we can infer that the impact from $v_{i}$ on $v_{j}$ is critically little.

\textbf{Reachability Score.} To calculate the correlation between traffic nodes across different time slots based on reachability knowledge, we novelly define the reachability score  as follows:

\begin{small}
\begin{equation}
a_{qp}^{ji}=\left\{\begin{array}{ll}1 & i=j\\
0 & p\delta + \text{M}_\text{ij} < (q-1)\delta, i\neq j\\
0 & p(\delta-1) + \text{M}_\text{ij} > q\delta, i\neq j\\
\frac{\text{M}_\text{ij}+p\delta-(q-1)\delta}{\delta} & (q-1)\delta  < p\delta + \text{M}_\text{ij},i\neq j\\
\frac{q\delta-\text{M}_\text{ij}-(p-1)\delta}{\delta} & q\delta  \geq p(\delta-1) + \text{M}_\text{ij},i\neq j\\
\end{array}\right.
\end{equation}
\end{small}where $[t_1, t_2]=[(p-1)\delta, p\delta]$, $[t_3, t_4]=[(q-1)\delta, q\delta]$. Here $\delta$ is time granularity, $p$ is the input step and $q$ is the output step. $a_{qp}^{ji} \in [0,1]$ is the attention score based on  reachability indicating the importance of context traffic node $i$ at step $p$ to target traffic node $j$ at step $q$. If there is little correlation between traffic condition $i$ and $j$, $a_{qp}^{ji}=0$ and the importance of a traffic node to itself is defined as $a_{qp}^{ii}=1$.

With the attention score $a_{qp}^{ji}$, we can construct the reachability matrix which represents the influence from the whole network at input step $p$ on the whole network at output step $q$, denoted as $A^{r}_{qp} \in \mathbb{R}^{\text{N} \times \text{N}}$ as follows:

\begin{equation}
\begin{split}
A^{r}_{qp}=
\left[\begin{array}{cccc}
{a_{qp}^{11}} & {...} &{a_{qp}^{1\text{N}}}\\
{...} & {...} &{...}\\
{a_{qp}^{\text{N}1}} & {...} &{a_{qp}^{\text{N}\text{N}}}\\
\end{array}\right]
\end{split}
\end{equation}

\textbf{Reachability Correlation Extraction.} Based on the reachability attention matrix, we can extract the spatial dependency from input step $p$ on output step $q$ with graph convolution network as follows:

\begin{equation}
\text{XR}_{qp} = \text{GCN}(A_{qp}^{r}, \text{XST}_{p})
\end{equation}
where $\text{XR}_{qp} \in \mathbb{R}^{\text{N} \times \text{F}_{\text{R}}}$ is the high-level features containing influence from step $p$ on step $q$. To measure the impact from all input steps on the output step $q$, we concatenate them together: $\text{XR}_{q} = [\text{XR}_{q1}, \dots,\text{XR}_{q \text{P}}] \in \mathbb{R}^{\text{N} \times \text{P}\text{F}_{\text{R}}}$.

We have extracted the spatial correlations by graph convolution network from three perspectives above. In the next section, we will extract the temporal dependency from the high-level features produced by GCN.

\subsection{\textbf{Multi-Head Temporal Attention Based Seq2Seq}}
In this paper, we aim to predict the traffic status of the traffic network over multiple future steps. This is a typical multi-step time series prediction problem. We employ the Sequence to Sequence (Seq2Seq)  model with an encoder and a decoder to extract the temporal dependency \cite{DBLP:conf/nips/SutskeverVL14}. Seq2Seq takes an input sequence to generate an output sequence with different length as follows:

\begin{small}
\begin{equation}
[\mathbf{X}_{1}, \cdots,\mathbf{X}_{p},\cdots, \mathbf{X}_{\mathbf{P}}]\stackrel{Seq2Seq}{\longrightarrow} [\hat{\mathbf{Y}}_{1}, \cdots,\hat{\mathbf{Y}}_{q},\cdots,\hat{\mathbf{Y}}_{\mathbf{Q}}]
\end{equation}
\end{small}

We leverage GRU (Gated Recurrent Units) \cite{chung2014empirical} as the encoder and decoder to learn the long-term temporal dependency in traffic data, for that GRU is a simple yet powerful and efficient variant of RNNs. 

\subsubsection{\textbf{Multi-Spatial Seq2Seq}}
We have captured three kinds of spatial dependencies through graph convolution networks respectively. Semantic matrix and adjacent matrix both contain the spatial correlations among traffic nodes at the same time slot while reachability matrix contains spatial dependency among traffic nodes at different time slots.  In other words, high-level features based on semantic correlation and adjacent correlation only contain spatial dependency extracted at each input step and we feed them into the encoder to learn the temporal dependency from all the input steps. However, the reachability correlation has already extracted the spatiotemporal dependency from all the previous input steps. Therefore, we directly feed the high-level features based on reachability into the decoder at each output step.  The Seq2Seq taking multi-spatial correlations as inputs is stated as follows:

\begin{equation}
\begin{split}
\text{H}_{p} &=\operatorname{GRU \_ Encoder}(W_1(\text{xf}_{p} \| \text{xa}_{p}), \text{H}_{p-1})\\
\text{C} &=\text{H}_{\text{P}}\\
\text{S}_{q} &=\operatorname{GRU \_ Decoder}(W_2(\text{C} \| \text{xr}_{q} \| \mathbf{Y}_{q-1}), \text{S}_{q-1})\\
\hat{\mathbf{Y}}_{q}&= \text{ReLu}(W_3\text{S}_{q})
\end{split}
\end{equation} where $\text{H}_{p}\in \mathbb{R}^{\text{F}_\text{H}}$ is the hidden state of encoder at input step $p$ and $C\in \mathbb{R}^{\text{F}_\text{H}}$ is the context vector, $S_{q}\in \mathbb{R}^{\text{F}_\text{S}}$ is the decoder hidden state at output step $q$.  $\mathbf{Y}_{q-1}\in \mathbb{R}^{\text{F}_\text{O}}$ is the ground truth at step $q-1$ and $\hat{\mathbf{Y}}_{q}\in \mathbb{R}^{\text{F}_\text{O}}$ is the prediction output at step $q$. $\{W_{1}, W_{2},W_{3}\}$ are the trainable parameters. $\text{xf}_{p} \in \mathbb{R}^{\text{F}_\text{F}}, \text{xa}_{p} \in \mathbb{R}^{\text{F}_\text{A}}, \text{xr}_{q} \in \mathbb{R}^{\text{P}\text{F}_\text{R}}$ are vectors from $\text{XF}_{p}, \text{XA}_{p}, \text{XR}_{q}$, representing high-level features of a traffic node based on semantic similarity, adjacent trend and reachability respectively. 

Following \cite{DBLP:conf/iclr/LiYS018}, we integrate scheduled sampling into Seq2Seq to ease the prediction error accumulation problem, which feeds $\mathbf{Y}_{q-1}$ into the model with $\epsilon_{i}$ probability and $\hat{\mathbf{Y}}_{q-1}$ with $1-\epsilon_{i}$ at $i_{th}$ iteration during training process.

\subsubsection{\textbf{Multi-Temporal Attention}} 
The previous traffic conditions at different steps of the encoder can impact each future step of the decoder differently \cite{DBLP:conf/iclr/LiYS018}. To model such relationship, a temporal attention allows the decoder to focus on relevant historical observations for any given future step by assigning different attention weights to every encoder hidden state $[\text{H}_{1},\cdots,\text{H}_{\text{P}}]$. Inspired by \cite{DBLP:conf/emnlp/LuongPM15}, we calculate the attention score between historical step $p$ and future step $q$  and normalize it through a Softmax layer:

\begin{small}
\begin{equation}
\begin{split}
e_{qp} &=(v)^{T}\text{tanh}(W[\text{H}_{p}\|\text{S}_{q-1}]) \\
\alpha_{qp}&=\frac{\exp (e_{qp})}{\sum_{i=1}^{\text{P}} \exp (e_{qi})}
\end{split}
\end{equation}
\end{small}where $W \in \mathbb{R}^{\text{F}_\text{e}\times (\text{F}_\text{H}+\text{F}_\text{S})}$ and $v \in \mathbb{R}^{\text{F}_\text{e}}$. With the temporal attention scores, the decoder hidden state $\text{S}_{q}$ can be updated as follows:
\begin{small}
\begin{equation}
\begin{split}
\text{C}_{q}&=\sum_{p=1}^{\text{P}} \alpha_{qp} \text{H}_{p}\\
\text{S}_{q} &=\operatorname{GRU \_ Decoder}(W_2(\text{C}_{q} \| \text{xr}_{q} \| \mathbf{Y}_{q-1}), \text{S}_{q-1})
\end{split}
\end{equation}
\end{small}To learn more complicated time dependency in different perspectives, we also extend the temporal attention to $\text{H}$ heads as follows:
\begin{small}
\begin{equation}
\begin{split}
e_{qp}^{h}&=(v_{h})^{T}\text{tanh}(W_{h}[\text{H}_{p}\|\text{S}_{q-1}])\\
\alpha_{qp}^{h}&=\frac{\exp (e_{qp}^{h})}{\sum_{i=1}^{\text{P}} \exp (e_{qi}^{h})}\\
\text{C}_{q}&=\sum_{p=1}^{\text{P}}W^{C}_{h}( \|_{h=1}^{\text{H}}\alpha_{qp}^{h}\text{H}_{p})
\end{split}
\end{equation}
\end{small}
where $W_{h}\in \mathbb{R}^{\text{F}_\text{e}\times (\text{F}_\text{H}+\text{F}_\text{S})} $ and $v_{h}\in \mathbb{R}^{\text{F}_\text{e}}$, $W^{C}_{h} \in \mathbb{R}^{\text{F}_\text{H} \times \text{H}\text{F}_\text{H}}$ .

\section{Experiments}
This paper focuses on the research questions as follows:

(1) How does our model MSGC-Seq2Seq perform at prediction accuracy and efficiency compared with other benchmarks on different datasets?

(2) Does each component in our model make contributions for prediction? 

(3) What's the performance of our model when the traffic data is insufficient and noisy?

(4) How does our model react to parameter sensitivity test?

To answer the research questions above, we conduct extensive experiments on two real-world highway traffic datasets.

\subsection{\textbf{Datasets}}
We compare our model with baselines on the following highway traffic datasets.

\textbf{METR} collects traffic speed by sensors on the highway of Los Angeles County. We utilize 207 sensors in the highway and the time period observed is from Mar 1st to Jun 30th in 2012, i.e. four months.

\textbf{PEMS} datasets are from California Transportation Agencies’ (CalTrans) Performance Measurement System (PeMS). 
There are more than 39,000 sensors deployed on the highway in the major metropolitan areas in California \cite{DBLP:conf/aaai/GuoLFSW19}. The Geographic information about each sensor is recorded in the datasets. The traffic measurement chosen in this experiment is the average traffic speed, which is collected in real time every 30 seconds and aggregated in 5 minutes. The dataset in this paper contains 6 months data ranging from Jan to May in 2017 in the San Francisco Bay Area and 325 sensors are selected.

%We choose three datasets from two areas in different periods, defined as PEMS-Ber 2016, PEMS-Bay 2017, PEMS-Bay 2018. PEMS-Ber 2016 is collected in San Bernardino from July to August in 2016 with 170 detectors.  PEMS-Bay 2017 contains 6 months data ranging from Jan to May in 2017 in the San Francisco Bay Area with 325 sensors selected in this network while PEMS-Bay 2018 contains one month data from January to February in 2018 with 307 detectors. 

For all the datasets, 70\% data is utilized for training, 10\% for validation and the rest 10\% for testing, and Z-Score normalization \cite{DBLP:conf/iclr/LiYS018} is applied to normalize the datasets. We follow the way in \cite{DBLP:conf/iclr/LiYS018} to build the adjacent matrix. The details of the datasets are shown in Table \ref{tab:detail}.

\begin{table}[htb]
	\setlength\tabcolsep{1.5pt}   %%设置列距离,默认6PT
	\centering
	\footnotesize
	\setlength{\abovecaptionskip}{2pt} % 设置图片与caption的距离
	\caption{The Details of Datasets}
	\begin{tabular}{lcccccc} %设置了每一列的宽度，强制转换。
		%\begin{tabular}{cp{0.5cm}p{0.5cm}p{0.5cm}p{0.5cm}p{0.5cm}p{0.5cm}p{0.5cm}p{0.5cm}p{0.5cm}p{0.5cm}}
		\toprule
		\textbf{Datasets}&\textbf{Date}&\textbf{Days}&\textbf{Sensors}&\textbf{Train}&\textbf{Val}&\textbf{Test}\\ 
		\midrule
		\textbf{METR}&03/01-06/27(2012)&119&207&4,966,012&709,430&1,418,860\\
		\textbf{PEMS}&01/01-06/31(2017)&181&325&11,859,120&1,694,160&3,388,320\\
		\cline{1-7}
	\end{tabular}
	\label{tab:detail}
\end{table}

\begin{table*}[htb]
	\centering
	\vspace{-0.35cm} %设置与上面正文的距离
	\setlength\tabcolsep{1pt}
	\setlength{\abovecaptionskip}{2pt}% 设置图片与caption的距离
	\caption{Evaluation of Various Approaches for Traffic State Prediction on Different Datasets}
	\footnotesize
	\begin{tabular}{c|c|c|c|c|c|c|c|c|c|c}
		\toprule
		\multirow{2}{*}{\textbf{\shortstack{Datasets}}} &\textbf{Steps}&\multicolumn{3}{c|}{\textbf{3 (15min)}}&\multicolumn{3}{c|}{\textbf{6 (30min)}}&\multicolumn{3}{c}{\textbf{12 (60min)}}\\
		\cline{2-11}
		&\textbf{Metrics}
		&\textbf{MAE}&\textbf{RMSE}&\textbf{MAPE (\%)}
		&\textbf{MAE}&\textbf{RMSE}&\textbf{MAPE (\%)}
		&\textbf{MAE}&\textbf{RMSE}&\textbf{MAPE (\%)}\\
		\midrule
		\multirow{9}{*}{\textbf{METR}}  
		&\scriptsize{HA}&\scriptsize{4.160±0.000}&\scriptsize{7.800±0.000}&\scriptsize{13.000±0.000}&\scriptsize{4.160±0.000}&\scriptsize{7.800±0.000}&\scriptsize{13.000±0.000}&\scriptsize{4.160±0.000}&\scriptsize{7.800±0.000}&\scriptsize{13.000±0.000}\\
		&\scriptsize{ARIMA}&\scriptsize{3.990±0.000}&\scriptsize{8.210±0.000}&\scriptsize{9.600±0.000}&\scriptsize{5.150±0.000}&\scriptsize{10.450±0.000}&\scriptsize{12.700±0.000}&\scriptsize{6.900±0.000}&\scriptsize{13.230±0.000}&\scriptsize{17.400±0.000}\\
		&\scriptsize{VAR}&\scriptsize{4.420±0.000}&\scriptsize{7.890±0.000}&\scriptsize{10.200±0.000}&\scriptsize{5.410±0.000}&\scriptsize{9.130±0.000}&\scriptsize{12.700±0.000}&\scriptsize{6.520±0.000}&\scriptsize{10.110±0.000}&\scriptsize{15.800±0.000}\\
		&\scriptsize{SVR}&\scriptsize{3.990±0.000}&\scriptsize{8.450±0.000}&\scriptsize{9.300±0.000}&\scriptsize{5.050±0.000}&\scriptsize{10.870±0.000}&\scriptsize{12.100±0.000}&\scriptsize{6.720±0.000}&\scriptsize{13.760±0.000}&\scriptsize{16.700±0.000}\\
		&\scriptsize{FNN}&\scriptsize{3.990±0.110}&\scriptsize{7.517±0.135}&\scriptsize{12.088±1.052}&\scriptsize{4.266±0.010}&\scriptsize{8.581±0.033}&\scriptsize{12.103±0.096}&\scriptsize{4.490±0.000}&\scriptsize{9.410±0.024}&\scriptsize{12.497±0.070}\\
		&\scriptsize{FC-LSTM}&\scriptsize{3.451±0.006}&\scriptsize{6.021±0.018}&\scriptsize{9.210±0.024}&\scriptsize{3.798±0.024}&\scriptsize{7.096±0.049}&\scriptsize{10.933±0.148}&\scriptsize{4.372±0.001}&\scriptsize{8.311±0.036}&\scriptsize{13.447±0.109}\\
		&\scriptsize{STGCN}&\scriptsize{2.821±0.001}&\scriptsize{5.353±0.008}&\scriptsize{7.227±0.009}&\scriptsize{3.366±0.006}&\scriptsize{6.567±0.016}&\scriptsize{9.250±0.037}&\scriptsize{3.981±0.019}&\scriptsize{7.808±0.013}&\scriptsize{11.733±0.145}\\
		&\scriptsize{DCRNN}&\scriptsize{2.696±0.015}&\scriptsize{4.994±0.005}&\scriptsize{6.780±0.008}&\scriptsize{3.092±0.065}&\scriptsize{5.993±0.181}&\scriptsize{8.150±0.144}&\scriptsize{3.452±0.114}&\scriptsize{6.904±0.235}&\scriptsize{9.597±0.388}\\
		&\scriptsize{GAMAN}&\scriptsize{2.646±0.005}&\scriptsize{5.045±0.020}&\scriptsize{6.725±0.055}&\scriptsize{2.934±0.006}&\scriptsize{5.810±0.022}&\scriptsize{7.953±0.121}&\scriptsize{3.341±0.005}&\scriptsize{6.818±0.054}&\scriptsize{9.783±0.213}\\
		&\scriptsize{MSGC-Seq2Seq}&\scriptsize{\textbf{2.503±0.011}}&\scriptsize{\textbf{4.732±0.018}}&\scriptsize{\textbf{6.247±0.061}}&\scriptsize{\textbf{2.757±0.003}}&\scriptsize{\textbf{5.470±0.027}}&\scriptsize{\textbf{7.230±0.050}}&\scriptsize{\textbf{3.088±0.010}}&\scriptsize{\textbf{6.457±0.053}}&\scriptsize{\textbf{8.700±0.078}}\\
		\cline{1-11}
		
		\multirow{9}{*}{\textbf{PEMS}}
		&\scriptsize{HA}&\scriptsize{2.880±0.000}&\scriptsize{5.590±0.000}&\scriptsize{6.800±0.000}&\scriptsize{2.880±0.000}&\scriptsize{5.590±0.000}&\scriptsize{6.800±0.000}&\scriptsize{2.880±0.000}&\scriptsize{5.590±0.000}&\scriptsize{6.800±0.000}\\
		&\scriptsize{ARIMA}&\scriptsize{1.620±0.000}&\scriptsize{3.300±0.000}&\scriptsize{3.500±0.000}&\scriptsize{2.330±0.000}&\scriptsize{4.760±0.000}&\scriptsize{5.400±0.000}&\scriptsize{3.380±0.000}&\scriptsize{6.500±0.000}&\scriptsize{8.300±0.000}\\
		&\scriptsize{VAR}&\scriptsize{1.740±0.000}&\scriptsize{3.160±0.000}&\scriptsize{3.600±0.000}&\scriptsize{2.320±0.000}&\scriptsize{4.250±0.000}&\scriptsize{5.000±0.000}&\scriptsize{2.930±0.000}&\scriptsize{5.440±0.000}&\scriptsize{6.500±0.000}\\
		&\scriptsize{SVR}&\scriptsize{1.850±0.000}&\scriptsize{3.590±0.000}&\scriptsize{3.800±0.000}&\scriptsize{2.480±0.000}&\scriptsize{5.180±0.000}&\scriptsize{5.500±0.000}&\scriptsize{3.280±0.000}&\scriptsize{7.080±0.000}&\scriptsize{8.000±0.000}\\
		&\scriptsize{FNN}&\scriptsize{2.149±0.090}&\scriptsize{4.613±0.856}&\scriptsize{5.327±0.948}&\scriptsize{2.291±0.046}&\scriptsize{5.077±0.501}&\scriptsize{5.613±0.528}&\scriptsize{2.481±0.006}&\scriptsize{5.684±0.061}&\scriptsize{6.167±0.167}\\
		&\scriptsize{FC-LSTM}&\scriptsize{2.196±0.003}&\scriptsize{4.498±0.018}&\scriptsize{5.045±0.039}&\scriptsize{2.320±0.013}&\scriptsize{4.809±0.014}&\scriptsize{5.382±0.018}&\scriptsize{2.451±0.003}&\scriptsize{5.125±0.008}&\scriptsize{5.813±0.029}\\
		&\scriptsize{STGCN}&\scriptsize{1.349±0.014}&\scriptsize{2.770±0.007}&\scriptsize{2.783±0.033}&\scriptsize{1.768±0.030}&\scriptsize{3.912±0.038}&\scriptsize{3.960±0.110}&\scriptsize{2.417±0.022}&\scriptsize{5.429±0.064}&\scriptsize{5.857±0.029}\\
		&\scriptsize{DCRNN}&\scriptsize{1.328±0.061}&\scriptsize{2.597±0.036}&\scriptsize{2.663±0.139}&\scriptsize{1.700±0.071}&\scriptsize{3.580±0.056}&\scriptsize{3.600±0.148}&\scriptsize{1.991±0.093}&\scriptsize{4.536±0.131}&\scriptsize{4.607±0.186}\\
		&\scriptsize{GAMAN}&\scriptsize{1.264±0.010}&\scriptsize{2.649±0.033}&\scriptsize{2.584±0.048}&\scriptsize{1.619±0.018}&\scriptsize{3.627±0.044}&\scriptsize{3.582±0.114}&\scriptsize{1.816±0.011}&\scriptsize{4.069±0.020}&\scriptsize{4.217±0.056}\\
		&\scriptsize{MSGC-Seq2Seq}&\scriptsize{\textbf{1.102±0.005}}&\scriptsize{\textbf{2.255±0.017}}&\scriptsize{\textbf{2.210±0.013}}&\scriptsize{\textbf{1.344±0.006}}&\scriptsize{\textbf{3.024±0.033}}&\scriptsize{\textbf{2.870±0.030}}&\scriptsize{\textbf{1.592±0.006}}&\scriptsize{\textbf{3.678±0.009}}&\scriptsize{\textbf{3.545±0.055}}\\
		\cline{1-11}
		\bottomrule
	\end{tabular}
	
	\label{tab:Results}
\end{table*}

\subsection{\textbf{Experimental Settings}}
\subsubsection{\textbf{Evaluation Metrics}}
MAE (Mean Absolute Error), RMSE (Root Mean Square Error) and MAPE (Mean Absolute Percentage Error) are the most widely utilized evaluation metrics in traffic state prediction tasks. Following previous works \cite{DBLP:conf/aaai/GuoLFSW19,DBLP:journals/corr/abs-1911-08415,DBLP:conf/iclr/LiYS018}, we utilize them to evaluate the model performance.

\subsubsection{\textbf{Baselines}}
We compare our approach with the following baselines, which can be categorized as traditional statistic methods (i.e. HA, VAR, ARIMA, SVR) and simple deep learning methods (i.e. FNN, FC-LSTM), graph-based deep learning methods (STGCN, DCRNN, GMAN).

\textbf{HA}: Historical Average, which predicts the future traffic conditions by averaging the traffic conditions at historical time.

\textbf{VAR}: Vector Auto-Regression is extensively leveraged in time series prediction task. It is a simple multivariate model where each
variable is explained by its own past values and the past values of all the other variables in the system. 

\textbf{ARIMA} \cite{williams2003modeling}: Auto-regressive integrated moving average, a well known model for predicting time series data. 

%\textbf{Lasso/Ridge}: Lasso is a linear regression method with L1 regularizer and can choose the main variables from a bunch of variables while Ridge is a linear regression method with  L2 regularizer which tends to assign weights to different input features evenly. We  utilize the MultiTaskLassoCV and RidgeCV class from Sklearn respectively to implement the linear functions.

\textbf{SVR}: Support Vector Regression is a regression method utilizing linear support vector machine in prediction.
 
\textbf{FNN}: Feed Foward Neural Network can extract the nonlinearity in traffic data.

\textbf{FC-LSTM} \cite{DBLP:conf/nips/SutskeverVL14}: A variant of Long Short Term Memory Network with fully connected layer as the output layer.

\textbf{STGCN} \cite{DBLP:conf/ijcai/YuYZ18}: Spatio-temporal graph convolutional network which consists of graph convolutional layers and 1D convolution layers and can capture both spatial and temporal dependencies in traffic prediction.

\textbf{DCRNN} \cite{DBLP:conf/iclr/LiYS018}: Diffusion Convolutional Recurrent Neural Network, which integrates diffusion convolution with recurrent neural network to capture the spatial and temporal properties in traffic data. 

\textbf{GAMAN} \cite{DBLP:journals/corr/abs-1911-08415}: Graph Multi-Attention Network, which designs an encoder-decoder architecture with multiple spatio-temporal attention mechanisms to predict traffic conditions. 

\subsubsection{\textbf{Parameter Settings}}
For baselines STGCN, DCRNN, GMAN, we utilize the parameter settings as suggested by the original papers as much as possible. Following previous works \cite{DBLP:conf/iclr/LiYS018}, the number of input steps is set the same as output steps. For instance, we use $\text{P}=3$ steps (15 minutes) to predict the next $\text{Q}=3$ steps (15 minutes). In terms of training, Adam optimizer \cite{kingma2014adam} is utilized to optimize all the deep learning methods. The maximum number of epochs is set as 1000 and the batch size is set as 16. The the initial learning rate is 0.001 with a learning rate decay strategy. Our model is developed using Python and TensorFlow 1.x. Most of the experiments are run on a GPU (32GB) machine with TESLA V100.

The hyper-parameters in our model include the spatiotemporal embedding dimension $\text{F}_{\text{ST}}$, the number of hidden units of GRU in encoder $\text{F}_{\text{H}}$ and decoder $\text{F}_{\text{S}}$, the number of heads in temporal attention $\text{H}$, the numbers of hidden units in adjacency GCN, semantic GCN and reachability GCN, i.e. $[\text{F}_{\text{A}}, \text{F}_{\text{F}}, \text{F}_{\text{R}}]$. For simplicity, both encoder and decoder have two layers and each layer shares the same number of hidden units. All GCN in this paper have two layers and each layer shares the same number of hidden units. All parameters are tuned on the validation set and we get the best model performance on the setting $\text{F}_{\text{A}}=\text{F}_{\text{F}}= \text{F}_{\text{R}}=64,\text{F}_{\text{ST}}= 256, \text{F}_{\text{H}}= \text{F}_{\text{S}}=64, \text{H}=5$.

\begin{figure}[htb]
 \centering
 \subfigure[METR]{
 \includegraphics[width=0.49\textwidth, height=0.13\textheight]{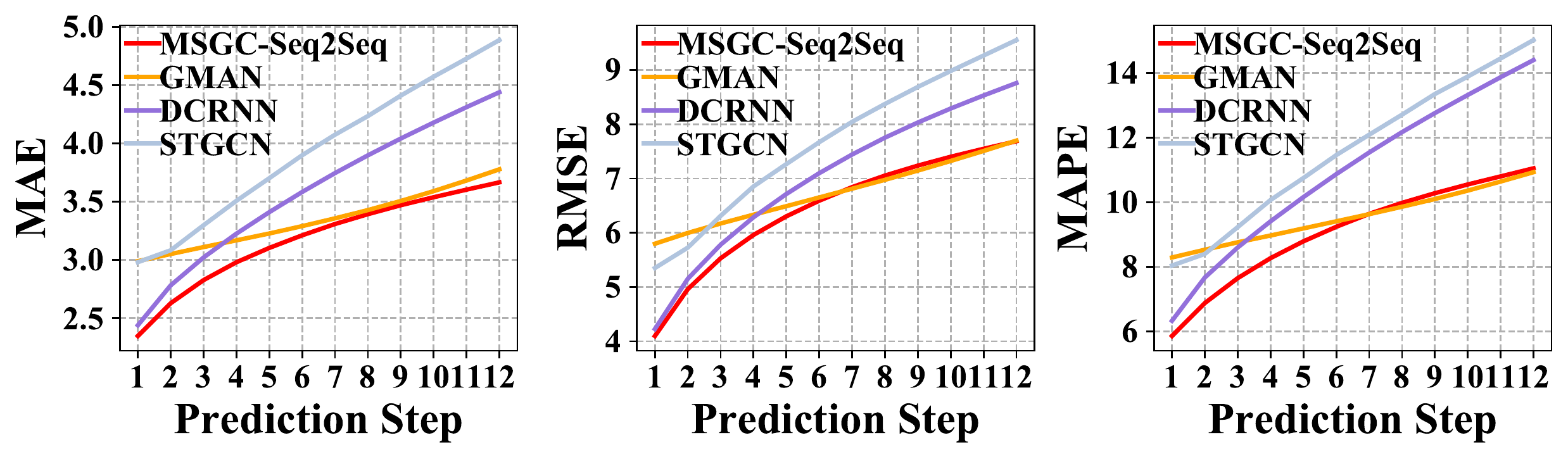}
 }\\
 \subfigure[PEMS]{
 \includegraphics[width=0.49\textwidth, height=0.13\textheight]{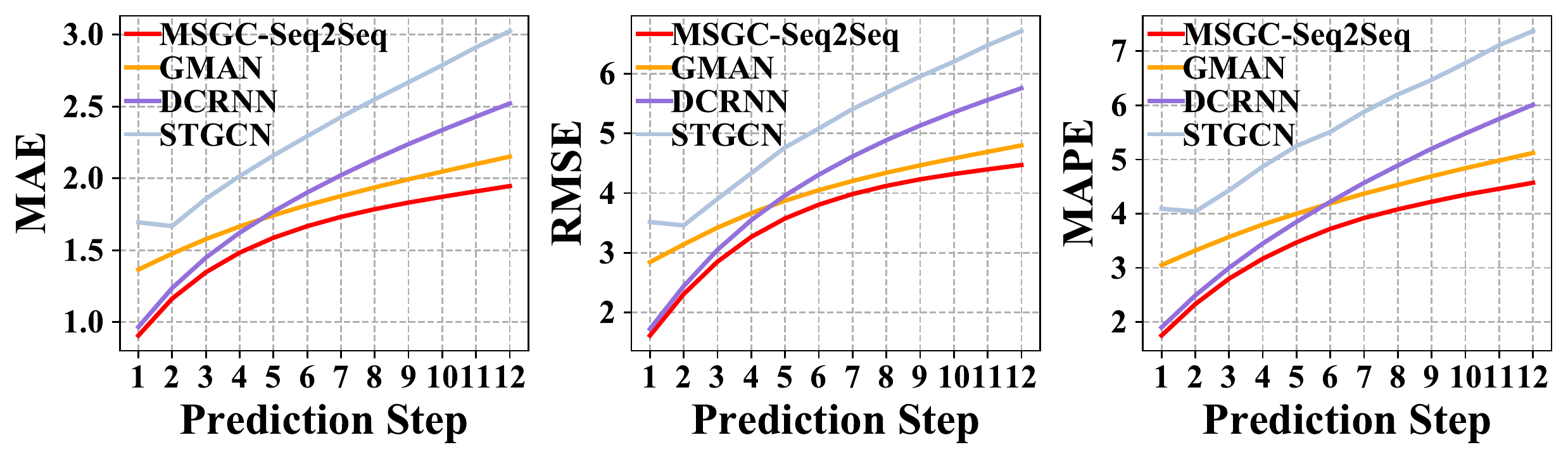}
 }
 \caption{Model Performances of Graph Deep Learning Methods at Each Output Step on Two Datasets For 12 Steps Prediction.}
 \label{fig:steps}
\end{figure}

\subsection{\textbf{Experimental Analysis}}
 To reduce the randomness of deep learning based methods in Table \ref{tab:Results}, we repeat the related experiments five times to get more  stable and  convincing results. We present their model performances with the means and standard errors of all metrics. In addition, the experiment results of traditional methods (i.e. HA, ARMIA,VAR, SVR) are from previous work \cite{DBLP:conf/iclr/LiYS018}. The best performance on each dataset is highlighted in bold font.

\subsubsection{\textbf{Approaches Comparison}} In this subsection, we compare our model with other benchmarks in the overall model performance, efficiency, and visualization.

\begin{figure*}[htb]
 \centering
 \includegraphics[width=0.9\textwidth, height=0.13\textheight]{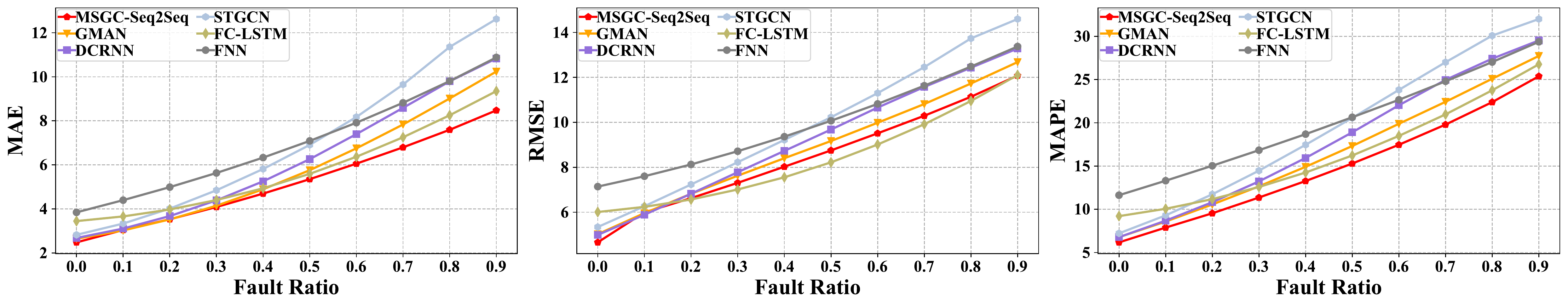}
 \caption{The Fault Tolerance Test of All the Deep  Learning Models on METR Dataset for 3 Steps with Fault Ratio Ranging from $[0\%, \cdots, 90\%]$}
 \label{fig:ft}
\end{figure*}

\begin{figure*}[htb]
 \centering
 \includegraphics[width=0.9\textwidth, height=0.13\textheight]{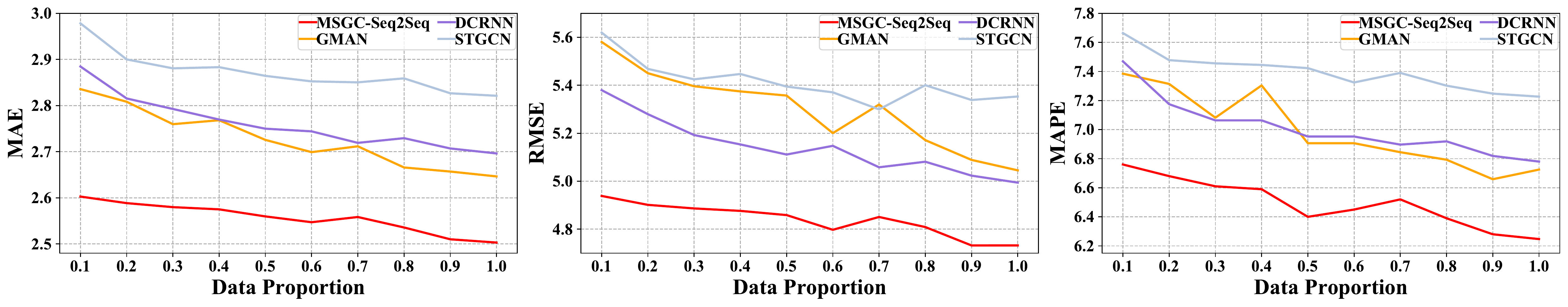}
 \caption{The Data Sparsity Test of Graph-based Deep Learning Models on METR Dataset for 3 Steps with Data Propotion Ranging from $[10\%, \cdots, 100\%]$}
 \label{fig:ds}
\end{figure*}

\begin{figure}[htb]
 %\centering
 \subfigure[Weekday]{
 \includegraphics[width=0.49\textwidth, height=0.13\textheight]{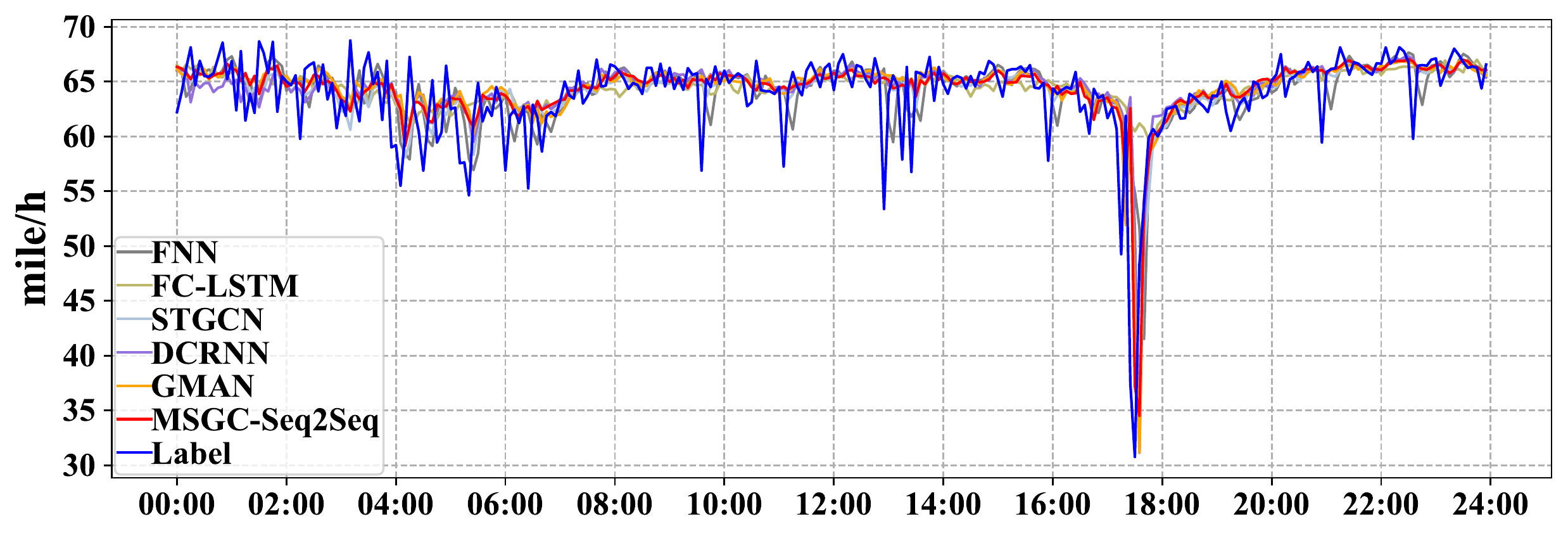}
 }\\
 \subfigure[Weekend]{
 \includegraphics[width=0.49\textwidth, height=0.13\textheight]{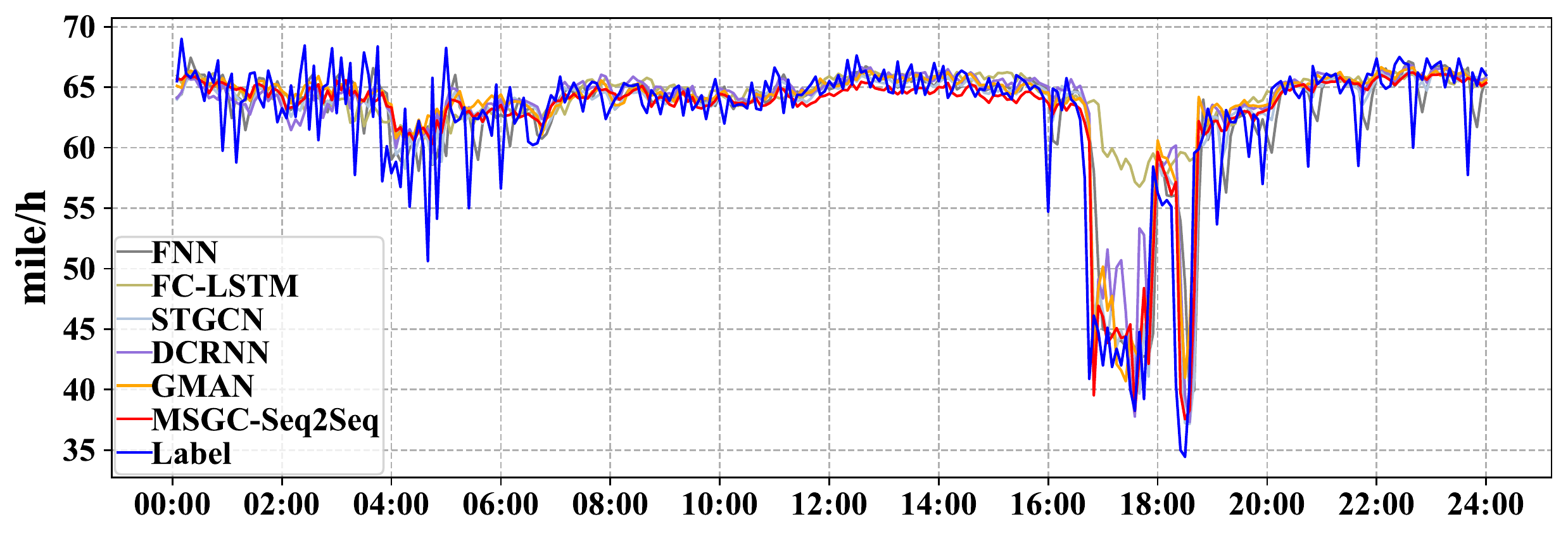}
 }
 \caption{The Comparison between Ground Truth and Predictions of All the Deep Learning Models on METR Dataset on One Day.}
 \label{fig:vis}
\end{figure}

\textbf{Model Performance.} As shown in Table \ref{tab:Results}, the graph-based deep learning methods (STGCN, DCRNN, GMAN, MSGC-Seq2Seq) perform significantly better than the simple deep learning methods (i.e. FNN, FC-LSTM) and statistic methods. It proves the effectiveness of extracting spatial correlation based on a traffic graph. When the predicted time period is short (i.e. 3 steps), some statistic method even has a better performance than the simple deep learning methods. For example, ARIMA performs better than FNN and FC-LSTM at PEMS dataset on 3 steps. However, as the output steps become longer, the deep learning methods show their superiority over statistic methods, probably because the spatiotemporal correlation become more complex in a long time prediction. Among the graph-based deep learning approaches, our model has the best performance on two datasets at all the metrics. On METR dataset, compared with STGCN, it has increased nearly 1\% in MAPE on 3 steps. When the predicted period become longer, the MAPE gap between these two models becomes larger, i.e. \textbf{2}\% on 6 steps, \textbf{3}\% on 12 steps. On PEMS dataset, the differences on MAPE between MSGC-Seq2Seq and STGCN become more obvious, i.e. 0.5\%, 1\%, \textbf{2.3}\% on 3, 6, 12 steps respectively. It proves that our model can handle long outsteps better than STGCN, perhaps due to its mechanisms to ease information dilution in a long term prediction.

\begin{table}[htbp]
 \centering
 \footnotesize
 \renewcommand\tabcolsep{2.0pt} % 调整表格列间的宽度
 \setlength{\abovecaptionskip}{2pt} 
 \caption{Efficiency Per Epoch of Deep Learning Models}
  \label{tab:eff}
 \begin{tabular}{ccccccc}
  \toprule
  \textbf{Models}&\textbf{FNN}&\textbf{FC-LSTM}&\textbf{STGCN}&\textbf{DCRNN}&\textbf{GMAN}&\textbf{MSGC-Seq2Seq}\\
  \midrule
\textbf{train}&15.4&19&57.8&54.5&123.7&70.7\\
 \textbf{test}&3.2&3.2&8.2&9.6&20&11.4\\
 \textbf{total}&19.5&23&68.1&66.6&147.4&84.9\\
  \cline{1-7}
 \end{tabular}
%\vspace{-0.5cm} 
\end{table}

\textbf{Multiple Prediction Steps Analysis.} We analyze the model performance at each output step of the graph-based deep learning methods which have a significantly better performance than other methods. As shown in Figure \ref{fig:steps}, nearly all models achieve the best performance at the first output step. As the time gap between input steps and output steps increases, the model performance become worse and worse. The deterioration rates of different models are different. The deterioration rate of DCRNN is the fastest while that of GMAN is the slowest. As to our model, it has a suboptimal deterioration rate.

\textbf{Efficiency.} Table \ref{tab:eff} shows the average time per epoch of all the deep learning based models on METR dataset for 3 steps prediction. FNN and FC-LSTM require the least time to finish an epoch, around 20 seconds for that they have the simplest structures and the fewest trainable parameters. Both STGCN and DCRNN need less than 60 seconds to finish training. GMAN has the longest time per epoch, nearly 400 seconds probably because it has the largest number of trainable parameters. Our model MSGC-Seq2Seq takes less than half of the time required by GMAN while it has the best performance on both datasets. Note that the reachability attention matrix in our model is calculated during the data preprocess.

\textbf{Visualization.} In order to visualize the prediction results of all the deep learning models, we randomly choose a node in the traffic network (i.e. a sensor) of METR dataset to observe the prediction values of its traffic speed on one day of weekday and weekend (also chosen randomly). Note that we choose the first output step prediction as the observed prediction values because the prediction at the first output step is the most accurate.  As shown in Figure \ref{fig:vis}, our model seems to have a better prediction at the peek hour in afternoon.

\subsubsection{\textbf{Model Ablation}}
\begin{table}[htbp]
\caption{The Degraded Model in METR Dataset (3 STEPS)} %显示表格的标题
\centering
\begin{tabular}{l|c|c|c} %设置了每一列的宽度，强制转换。
\hline
\textbf{Model} &\textbf{MAE}  &\textbf{RMSE}  &\textbf{MAPE}\\ %用&来分隔单元格的内容 \\表示进入下一行
\hline
\footnotesize{No Temporal Embedding}&\footnotesize{2.57±0.007}&\footnotesize{4.84±0.006}&\footnotesize{6.46±0.030}\\
\footnotesize{No Spatial Embedding}&\footnotesize{2.74±0.003}&\footnotesize{5.12±0.008}&\footnotesize{6.91±0.030}\\
\footnotesize{No Spatiotemporal Embedding}&\footnotesize{2.74±0.001}&\footnotesize{5.11±0.004}&\footnotesize{6.84±0.010}\\
\footnotesize{No Adjacent Correlation}&\footnotesize{2.52±0.002}&\footnotesize{\textbf{4.71±0.010}}&\footnotesize{6.28±0.015}\\
\footnotesize{No Semantic Correlation}&\footnotesize{2.70±0.001}&\footnotesize{5.08±0.011}&\footnotesize{6.81±0.010}\\
\footnotesize{No Reachability Correlation}&\footnotesize{2.54±0.021}&\footnotesize{4.78±0.052}&\footnotesize{6.34±0.110}\\
\footnotesize{No Temporal Attention}&\footnotesize{2.56±0.007}&\footnotesize{4.79±0.007}&\footnotesize{6.41±0.010}\\
\footnotesize{MSGC-Seq2Seq}&\footnotesize{\textbf{2.50±0.011}}&\footnotesize{4.73±0.020}&\footnotesize{\textbf{6.24±0.064}}\\
\hline
\end{tabular}
\label{tab:Ablation}
\end{table}

We integrate spatial attribute embedding, temporal attribute embedding, three kinds of spatial correlations and multi-head temporal attention mechanism into our model MSGC-Seq2Seq. To evaluate the effect of these mechanisms, we remove them from the original model to observe the performance of the corresponding degraded models. We conduct the experiments on METR dataset for 3 steps prediction. The results in Table \ref{tab:Ablation} show that the prediction performance decreases no matter which mechanism to remove, indicating that all mechanisms make contributions to improve the model performance. However, as we can see in Table \ref{tab:Ablation}, their contributions are different. When the Spatiotemporal Embedding or the Semantic Correlation are removed, the model performance becomes worst, referring that they play more important roles than other mechanisms. 

%While we remove the Adjacent Correlation or Reachability Correlation or Temporal Attention, the model performance become worse slightly indicating they are not very efficient in short predict period, i.e. 3 steps with 15 minutes.

\subsubsection{\textbf{Fault-tolerance Test}}
In real world traffic scenarios, we might collect the low-quality data with missing or wrong values due to the limited data collection methods and tools. In addition, mistakes brought by some data processing methods also decrease the data reliability. Therefore, the robustness of an approach to process noisy data is important for real world application. In this subsection, we want to test the effectiveness of our model and compare it with other benchmarks. We choose a set of fault-ratio from 10\% to 90\% to replace the traffic data values with noise of zero values. As we can see from Figure \ref{fig:ft}, the performances of all models on all metrics deteriorate quickly as the proportion of noise increases, however, in different speed. Among the graph deep learning methods, performance of STGCN becomes worse more quickly than others which refers that its robustness is the worst. Surprisingly, FC-LSTM has the lowest rate of deterioration among all the deep learning methods. When the fault-ratio is 90\%, FC-LSTM has the suboptimal performance while STGCN has the worst performance. The relative performance of ANN also becomes better as the data becomes noisier. This implicates that simple deep learning models have better capacity to combat noise. However, our model also has a slower deteriorated rate and its performance evaluated by MAE and MAPE are the best on all the fault-ratio settings. This proves that
\textbf{MSGC-Seq2Seq is effective and robust in the cases where the data is noisy}. The mechanisms we design to enhance the spatiotemporal information in the model might play a key role against noisy data.
 
\subsubsection{\textbf{Data Sparsity Test}}We want to test the effectiveness of graph deep learning models in sparse data scene.We randomly sample several subdatasets from METR at a proportion ranging from  10\% to 100\% and conduct experiments for 3 steps prediction. As shown in Figure \ref{fig:ds}, as the data scale decreases, most models generally have worse performances but in different speeds. The performance of our model decreases steadily while GMAN performance has a slight fluctuation. Note that our model performs better than other methods in all data scale, proving the effectiveness of our model and such effectiveness is more obvious when data size is smaller.

% Figure
\begin{figure} [htb]
\centering
\subfigcapskip=-8pt %ÉèÖÃ×ÓÍŒÓë×Ó±êÌâÖ®ŒäµÄŸàÀë

\subfigure[Spatiotemporal Embedding Dimension]{
\includegraphics[width=0.49\textwidth, height=0.10\textheight]{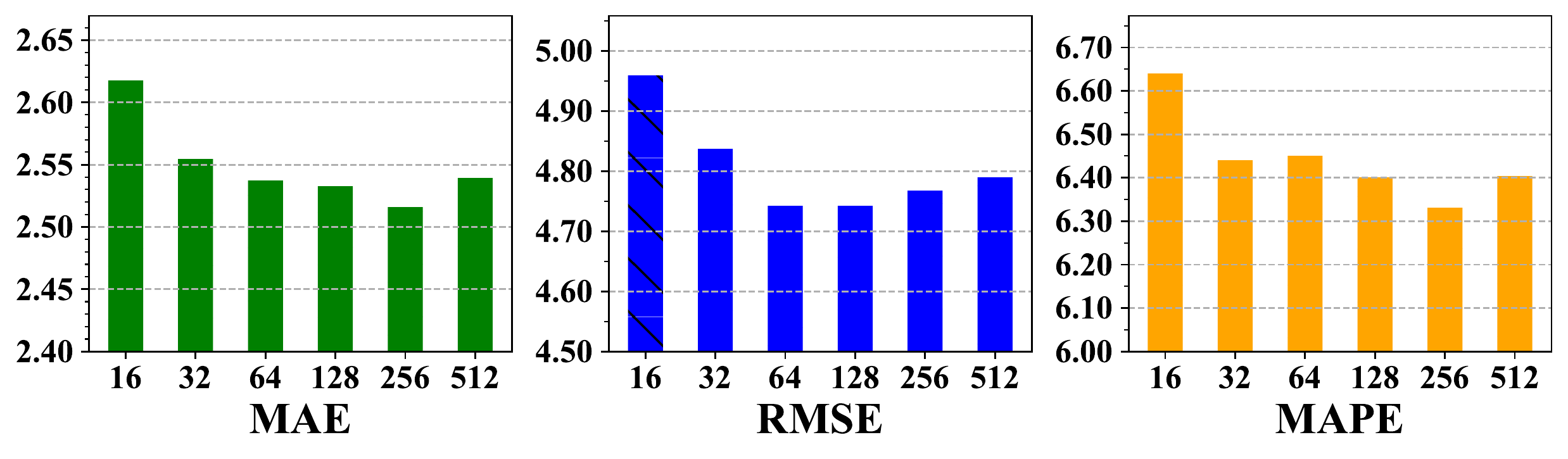}
}

\subfigure[The Number of GCN Units]{
\includegraphics[width=0.49\textwidth, height=0.10\textheight]{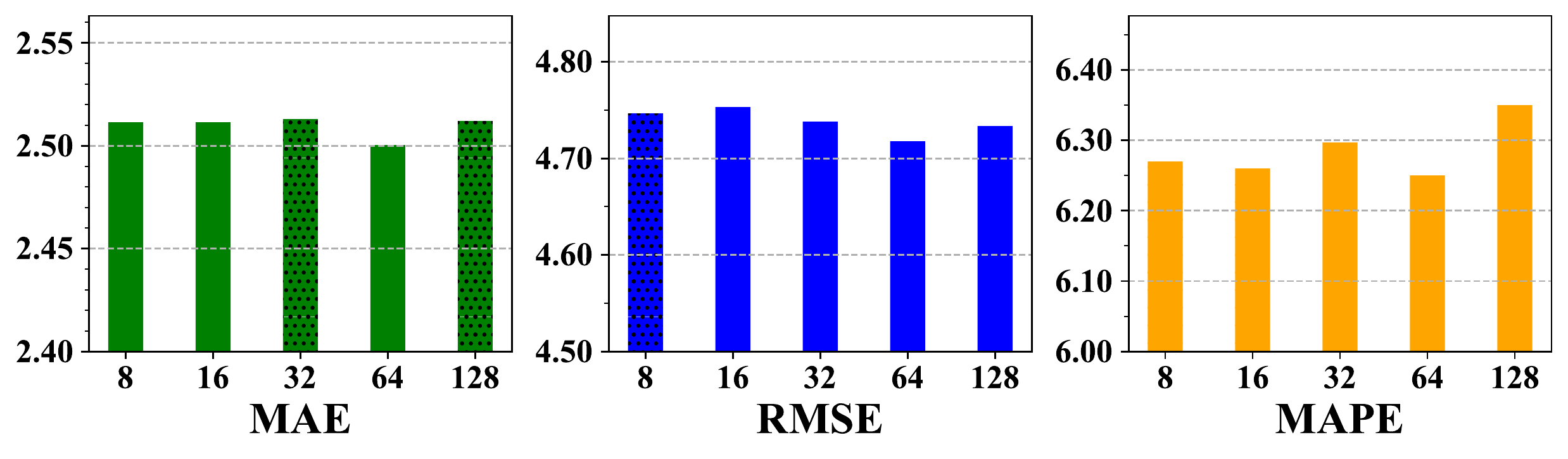}
}

\subfigure[The Number of GRU Units]{
\includegraphics[width=0.49\textwidth, height=0.10\textheight]{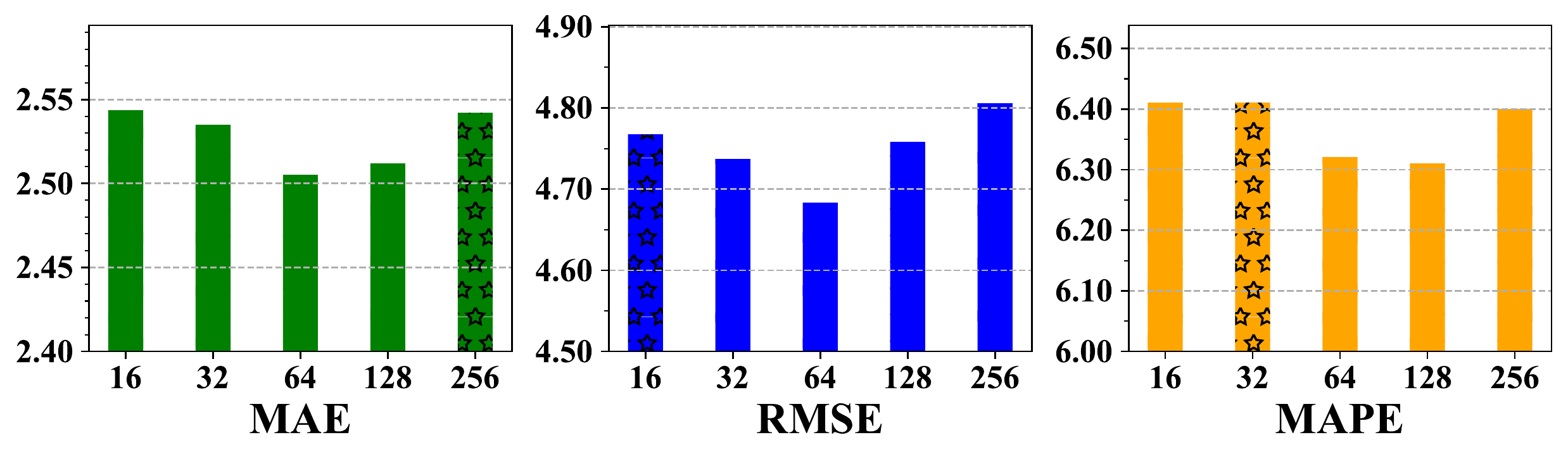}
}

\subfigure[The Number of Heads]{
\includegraphics[width=0.49\textwidth, height=0.10\textheight]{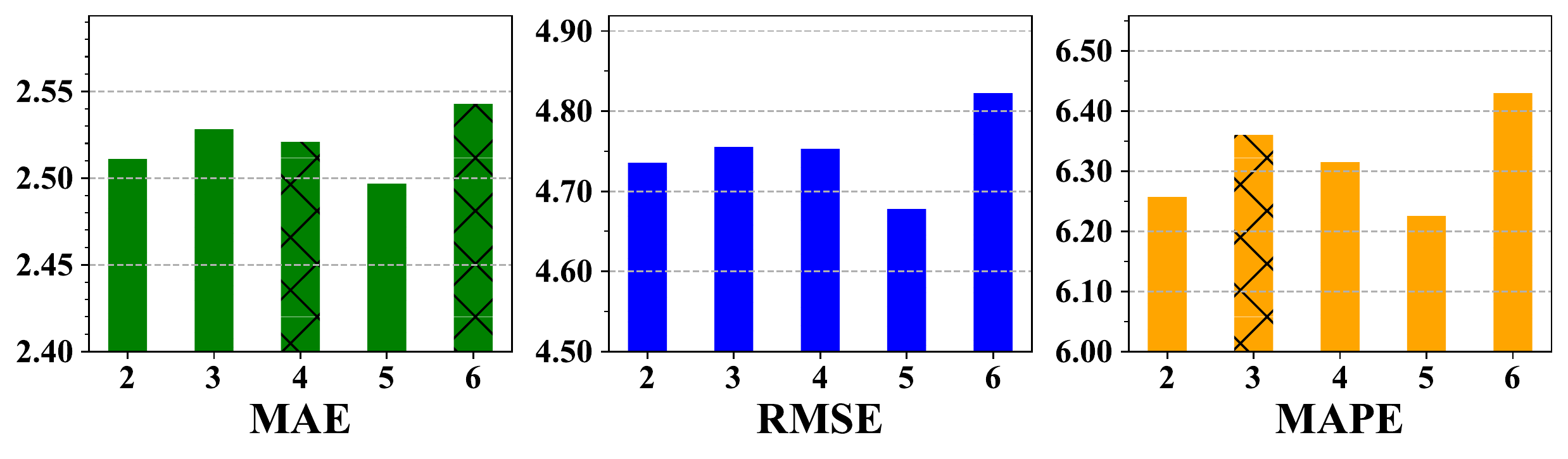}
}

\setlength{\abovecaptionskip}{-3pt} %±êÌâÖ®ÉÏ
\caption{Sensitivity Test on Different Hyper Parameters on METR Dataset for 3 Steps.}
\label{fig:sen}
%\vspace{-0.5cm} % ÍŒÆ¬žúÏÂÎÄµÄŸàÀë
\end{figure}

\subsubsection{\textbf{Parameter Sensitivity Analysis}}
In this section, we choose several important hyper-parameters to analyze the parameter sensitivity of our model on METR dataset for 3 steps prediction as follows:

\textbf{Spatiotemporal Embedding Dimension.} We conduct experiments on different embedding dimensions, i.e. [16, 32, 64, 128, 256, 512]. As shown in Figure \ref{fig:sen}, when the number of dimension units increases, the MAE decreases. When the dimension is 256, our model has the best performance at MAE. When it keeps increasing, the model performance decreases, perhaps due to overfitting problem.

\textbf{GCN Units.} We choose [8, 16, 32, 64, 128] GCN units to conduct the experiments. The smallest MAE is achieved at 64 units.

\textbf{GRU Units.} [16, 32, 64, 128, 256] GRU units are chosen to conduct the experiments. The smallest MAE is also achieved at 64 units.

\textbf{The Multiple Heads.} The head list test in the experiment is [1, 2, 3, 4, 5]. Figure \ref{fig:sen} shows that the best performance achieves at 5 heads.  If there are only 1 or 2 heads, the model might be unable to capture more complex correlation in traffic data. However, more heads might lead to more parameters, resulting in overfitting.

 \section{Related Work}
Traffic prediction has been extensively researched for many years. Compared with statistic methods in early stage (e.g. ARIMA \cite{DBLP:conf/icassp/YuZ04}, VAR \cite{DBLP:journals/jits/ChandraA09}, Kalman filtering \cite{DBLP:journals/cacie/XieZY07}), which can only capture linear correlation in traffic data, and traditional machine learning (e.g. Support Vector Machine \cite{DBLP:journals/ijon/FuMLL16}, K-Nearest Neighbors \cite{DBLP:conf/icdm/MayHKSS08}) which requires handcrafted feature engineering, deep learning based methods provide an end-to-end learning and have the superior capacity to capture the complex traffic pattern. The main ideas of most deep learning frameworks in traffic tasks can be summarized as: first aggregate the spatial dependencies between nodes at the same input step, then treat the aggregated traffic features over multiple input steps as a high-dimensional time series and extract the temporal dependency in the time series. 

To capture the spatial dependency, CNNs \cite{DBLP:conf/aaai/ZhangZQ17} decompose the network into grids while many traffic networks are graph based naturally. Recently, GNNs \cite{DBLP:conf/aaai/GuoLFSW19,DBLP:conf/aaai/Diao0ZLXH19} are used to aggregate features from neighbors based on the adjacent matrix or data-driven matrix \cite{DBLP:conf/aaai/GuoLFSW19,DBLP:conf/aaai/Diao0ZLXH19}. These models have extracted the spatial information of the network in each input step and achieved better performance than previous works. However, they can not work well to extract complex spatio-temporal dependencies in multi-step forecasting for that the spatial correlation between nodes across different time steps is  not well considered.

To extract the temporal dependency, LSTM and GRU are often adopted for one step prediction \cite{DBLP:conf/sdm/YuLSDL17, cuitraffic19,DBLP:conf/aaai/ChenLTZWWZ19}. For multi-step forecasting, Seq2Seq model with an encoder and a decoder is commonly utilized \cite{chen2019multi,DBLP:conf/iclr/LiYS018}. To distinguish the different impact of each input step on a given future step, a temporal attention mechanism, which can help a model make decisions by focusing on the important parts of input data \cite{DBLP:journals/corr/BahdanauCB14, DBLP:conf/aaai/ShenZLJPZ18, DBLP:conf/nips/MnihHGK14, DBLP:conf/coling/FengHYZ16}, is added to the decoder \cite{DBLP:conf/ijcai/BaiYK0S19, zhang2019multistep, DBLP:journals/corr/abs-1911-08415, DBLP:conf/aaai/GuoLFSW19}. However, few works pay attention to differentiate the contribution of different feature dimensions for prediction, which is important to help a model focus on the important input features.

\section{Conclusion}
In this paper, we focus on traffic network level multi-step traffic prediction. Compared with one-step traffic prediction, multi-step traffic prediction is more challenging because it requires to fully model the dynamic correlations among traffic nodes across time and space in a longer period. To address this problem, we design a novel deep learning model called MSGC-Seq2Seq. We first use the traffic features and
their spatiotemporal attributes via graph embedding methods to enrich the spatiotemporal information of model input. Afterward, we extract spatial correlations based on both data-driven knowledge (i.e. semantic similarity) and prior knowledge (geographical proximity and feature
similarity). The former can dig out hidden traffic patterns while the latter can provide more valuable spatial information, especially in insufficient or noisy data cases. Then we extract the temporal dependency by utilizing a GRU-based Seq2Seq model. We novelly develop a cross-step attention mechanism based on reachability to ease the dilution problem in Seq2Seq. In addition, we employ a multi-head temporal attention to distinguish the impact from different historical steps on each future step. We conduct experiments on two real-world traffic datasets. The experiments demonstrate that our model outperforms other baselines.

%\section*{Acknowledgment}
%The authors would like to thank anonymous reviewers for their valuable comments.
%
%This work is supported by the National Key R\&D Program of China (No.2019YFB2102100), National Natural Science Foundation of China (No.61802387), China’s Post-doctoral Science Fund (No.2019M663183), National Natural Science Foundation of Shenzhen (No.JCYJ20190812153212464), Shenzhen Engineering Research Center for Beidou Positioning Service Improvement Technology  (No.XMHT20190101035), Science and Technology Development Fund of Macao S.A.R (FDCT) under number 0015/2019/AKP, Shenzhen Discipline Construction Project for Urban Computing and Data Intelligence.

\bibliographystyle{IEEEtran}
\bibliography{IEEEabrv,reference}

% Generated by IEEEtran.bst, version: 1.12 (2007/01/11)
\begin{thebibliography}{10}
\providecommand{\url}[1]{#1}
\csname url@samestyle\endcsname
\providecommand{\newblock}{\relax}
\providecommand{\bibinfo}[2]{#2}
\providecommand{\BIBentrySTDinterwordspacing}{\spaceskip=0pt\relax}
\providecommand{\BIBentryALTinterwordstretchfactor}{4}
\providecommand{\BIBentryALTinterwordspacing}{\spaceskip=\fontdimen2\font plus
\BIBentryALTinterwordstretchfactor\fontdimen3\font minus
  \fontdimen4\font\relax}
\providecommand{\BIBforeignlanguage}[2]{{%
\expandafter\ifx\csname l@#1\endcsname\relax
\typeout{** WARNING: IEEEtran.bst: No hyphenation pattern has been}%
\typeout{** loaded for the language `#1'. Using the pattern for}%
\typeout{** the default language instead.}%
\else
\language=\csname l@#1\endcsname
\fi
#2}}
\providecommand{\BIBdecl}{\relax}
\BIBdecl

\bibitem{xie2020urban}
P.~Xie, T.~Li, J.~Liu, S.~Du, X.~Yang, and J.~Zhang, ``Urban flow prediction
  from spatiotemporal data using machine learning: A survey,''
  \emph{Information Fusion}, 2020.

\bibitem{ma2015long}
X.~Ma, Z.~Tao, Y.~Wang, H.~Yu, and Y.~Wang, ``Long short-term memory neural
  network for traffic speed prediction using remote microwave sensor data,''
  \emph{Transportation Research Part C: Emerging Technologies}, vol.~54, pp.
  187--197, 2015.

\bibitem{wang2018will}
D.~Wang, J.~Zhang, W.~Cao, J.~Li, and Y.~Zheng, ``When will you arrive?
  estimating travel time based on deep neural networks,'' in \emph{AAAI}, 2018.

\bibitem{9310691}
J.~Ye, J.~Zhao, K.~Ye, and C.~Xu, ``How to build a graph-based deep learning
  architecture in traffic domain: A survey,'' \emph{IEEE Transactions on
  Intelligent Transportation Systems}, pp. 1--21, 2020.

\bibitem{DBLP:conf/aaai/GuoLFSW19}
S.~Guo, Y.~Lin, N.~Feng, C.~Song, and H.~Wan, ``Attention based
  spatial-temporal graph convolutional networks for traffic flow forecasting,''
  in \emph{AAAI}, 2019, pp. 922--929.

\bibitem{DBLP:conf/iclr/LiYS018}
Y.~Li, R.~Yu, C.~Shahabi, and Y.~Liu, ``Diffusion convolutional recurrent
  neural network: Data-driven traffic forecasting,'' in \emph{{ICLR}}, 2018.

\bibitem{DBLP:journals/corr/abs-1906-00560}
X.~Zhou, Y.~Shen, and L.~Huang, ``Revisiting flow information for traffic
  prediction,'' \emph{arXiv:1906.00560}, 2019.

\bibitem{DBLP:conf/mdm/GeLLZ19}
L.~Ge, H.~Li, J.~Liu, and A.~Zhou, ``Temporal graph convolutional networks for
  traffic speed prediction considering external factors,'' in \emph{{MDM}},
  2019, pp. 234--242.

\bibitem{DBLP:journals/corr/abs-1903-07789}
J.~Sun, J.~Zhang, Q.~Li, X.~Yi, and Y.~Zheng, ``Predicting citywide crowd flows
  in irregular regions using multi-view graph convolutional networks,''
  \emph{IEEE Transactions on Knowledge and Data Engineering}, 2020.

\bibitem{DBLP:conf/ijcai/FangZMXP19}
S.~Fang, Q.~Zhang, G.~Meng, S.~Xiang, and C.~Pan, ``Gstnet: Global
  spatial-temporal network for traffic flow prediction,'' in \emph{Proceedings
  of the Twenty-Eighth International Joint Conference on Artificial
  Intelligence, {IJCAI} 2019}, 2019, pp. 2286--2293.

\bibitem{DBLP:journals/tits/YuG19}
J.~J.~Q. Yu and J.~Gu, ``Real-time traffic speed estimation with graph
  convolutional generative autoencoder,'' \emph{IEEE Transactions on
  Intelligent Transportation Systems}, vol.~20, no.~10, pp. 3940--3951, 2019.

\bibitem{DBLP:conf/aaai/Diao0ZLXH19}
Z.~Diao, X.~Wang, D.~Zhang, Y.~Liu, K.~Xie, and S.~He, ``Dynamic
  spatial-temporal graph convolutional neural networks for traffic
  forecasting,'' in \emph{AAAI}, 2019, pp. 890--897.

\bibitem{DBLP:journals/corr/abs-1903-05631}
B.~Yu, H.~Yin, and Z.~Zhu, ``St-unet: {A} spatio-temporal u-network for
  graph-structured time series modeling,'' \emph{arXiv:1903.05631}, 2019.

\bibitem{DBLP:conf/aaai/ChenLTZWWZ19}
C.~Chen, K.~Li, S.~G. Teo, X.~Zou, K.~Wang, J.~Wang, and Z.~Zeng, ``Gated
  residual recurrent graph neural networks for traffic prediction,'' in
  \emph{AAAI}, 2019, pp. 485--492.

\bibitem{DBLP:conf/kdd/LiHCSWZP19}
J.~Li, Z.~Han, H.~Cheng, J.~Su, P.~Wang, J.~Zhang, and L.~Pan, ``Predicting
  path failure in time-evolving graphs,'' in \emph{KDD}, 2019, pp. 1279--1289.

\bibitem{DBLP:conf/ijcai/WuPLJZ19}
Z.~Wu, S.~Pan, G.~Long, J.~Jiang, and C.~Zhang, ``Graph wavenet for deep
  spatial-temporal graph modeling,'' in \emph{IJCAI}, 2019.

\bibitem{DBLP:conf/ijcai/YuYZ18}
B.~Yu, H.~Yin, and Z.~Zhu, ``Spatio-temporal graph convolutional networks: {A}
  deep learning framework for traffic forecasting,'' in \emph{IJCAI}, 2018, pp.
  3634--3640.

\bibitem{DBLP:conf/ijcnn/ZhangWCC19}
Y.~Zhang, S.~Wang, B.~Chen, and J.~Cao, ``{GCGAN:} generative adversarial nets
  with graph {CNN} for network-scale traffic prediction,'' in \emph{IJCNN},
  2019, pp. 1--8.

\bibitem{DBLP:journals/corr/abs-1911-08415}
C.~Zheng, X.~Fan, C.~Wang, and J.~Qi, ``Gman: A graph multi-attention network
  for traffic prediction,'' in \emph{AAAI}, 2020.

\bibitem{zhang2019multistep}
Z.~Zhang, M.~Li, X.~Lin, Y.~Wang, and F.~He, ``Multistep speed prediction on
  traffic networks: A deep learning approach considering spatio-temporal
  dependencies,'' \emph{Transportation Research Part C: Emerging Technologies},
  vol. 105, pp. 297--322, 2019.

\bibitem{geng2019spatiotemporal}
X.~Geng, Y.~Li, L.~Wang, L.~Zhang, Q.~Yang, J.~Ye, and Y.~Liu, ``Spatiotemporal
  multi-graph convolution network for ride-hailing demand forecasting,'' in
  \emph{AAAI}, vol.~33, 2019, pp. 3656--3663.

\bibitem{guooptimized20}
K.~Guo, Y.~Hu, Z.~Qian, H.~Liu, and e.~Zhang, ``Optimized graph convolution
  recurrent neural network for traffic prediction,'' \emph{IEEE Transactions on
  Intelligent Transportation Systems}, pp. 1--12, 2020.

\bibitem{DBLP:conf/icde/HuG0J19}
J.~Hu, C.~Guo, B.~Yang, and C.~S. Jensen, ``Stochastic weight completion for
  road networks using graph convolutional networks,'' in \emph{{ICDE}}, 2019,
  pp. 1274--1285.

\bibitem{DBLP:conf/uai/ZhangSXMKY18}
J.~Zhang, X.~Shi, J.~Xie, H.~Ma, I.~King, and D.~Yeung, ``Gaan: Gated attention
  networks for learning on large and spatiotemporal graphs,'' in \emph{UAI},
  2018, pp. 339--349.

\bibitem{DBLP:journals/corr/abs-1903-00919}
B.~Yu, M.~Li, J.~Zhang, and Z.~Zhu, ``3d graph convolutional networks with
  temporal graphs: {A} spatial information free framework for traffic
  forecasting,'' \emph{arXiv:1903.00919}, 2019.

\bibitem{DBLP:journals/access/ZhangYL19a}
C.~Zhang, J.~J.~Q. Yu, and Y.~Liu, ``Spatial-temporal graph attention networks:
  {A} deep learning approach for traffic forecasting,'' \emph{IEEE Access},
  vol.~7, pp. 166\,246--166\,256, 2019.

\bibitem{9207049}
J.~{Ye}, J.~{Zhao}, K.~{Ye}, and C.~{Xu}, ``Multi-stgcnet: A graph convolution
  based spatial-temporal framework for subway passenger flow forecasting,'' in
  \emph{2020 International Joint Conference on Neural Networks (IJCNN)}, 2020,
  pp. 1--8.

\bibitem{vaswani2017attention}
A.~Vaswani, N.~Shazeer, N.~Parmar, J.~Uszkoreit, L.~Jones, A.~N. Gomez,
  L.~Kaiser, and I.~Polosukhin, ``Attention is all you need,'' \emph{arXiv
  preprint arXiv:1706.03762}, 2017.

\bibitem{DBLP:conf/kdd/GroverL16}
A.~Grover and J.~Leskovec, ``node2vec: Scalable feature learning for
  networks,'' in \emph{Proceedings of the 22nd {ACM} {SIGKDD} International
  Conference on Knowledge Discovery and Data Mining, 2016}, 2016, pp. 855--864.

\bibitem{DBLP:conf/kdd/PerozziAS14}
B.~Perozzi, R.~Al{-}Rfou, and S.~Skiena, ``Deepwalk: online learning of social
  representations,'' in \emph{The 20th {ACM} {SIGKDD} International Conference
  on Knowledge Discovery and Data Mining 2014}, 2014, pp. 701--710.

\bibitem{DBLP:journals/corr/BrunaZSL13}
J.~Bruna, W.~Zaremba, A.~Szlam, and Y.~LeCun, ``Spectral networks and locally
  connected networks on graphs,'' in \emph{ICLR}, 2014.

\bibitem{DBLP:conf/nips/DefferrardBV16}
M.~Defferrard, X.~Bresson, and P.~Vandergheynst, ``Convolutional neural
  networks on graphs with fast localized spectral filtering,'' in \emph{NIPS},
  2016, pp. 3837--3845.

\bibitem{DBLP:conf/iclr/KipfW17}
T.~N. Kipf and M.~Welling, ``Semi-supervised classification with graph
  convolutional networks,'' in \emph{ICLR}, 2017.

\bibitem{DBLP:conf/icpr/ZhangJCXP18}
Q.~Zhang, Q.~Jin, J.~Chang, S.~Xiang, and C.~Pan, ``Kernel-weighted graph
  convolutional network: {A} deep learning approach for traffic forecasting,''
  in \emph{{ICPR}}, 2018, pp. 1018--1023.

\bibitem{DBLP:conf/uic/LiPLXDMWB18}
J.~Li, H.~Peng, L.~Liu, G.~Xiong, B.~Du, H.~Ma, L.~Wang, and M.~Z.~A. Bhuiyan,
  ``Graph cnns for urban traffic passenger flows prediction,'' in
  \emph{SmartWorld}, 2018, pp. 29--36.

\bibitem{DBLP:conf/icde/HuLBCF16}
H.~Hu, G.~Li, Z.~Bao, Y.~Cui, and J.~Feng, ``Crowdsourcing-based real-time
  urban traffic speed estimation: From trends to speeds,'' in \emph{32nd {IEEE}
  International Conference on Data Engineering, {ICDE} 2016}, 2016, pp.
  883--894.

\bibitem{DBLP:conf/nips/SutskeverVL14}
I.~Sutskever, O.~Vinyals, and Q.~V. Le, ``Sequence to sequence learning with
  neural networks,'' in \emph{NIPS}, 2014, pp. 3104--3112.

\bibitem{chung2014empirical}
J.~Chung, C.~Gulcehre, K.~Cho, and Y.~Bengio, ``Empirical evaluation of gated
  recurrent neural networks on sequence modeling,'' in \emph{NIPS Workshop},
  2014.

\bibitem{DBLP:conf/emnlp/LuongPM15}
T.~Luong, H.~Pham, and C.~D. Manning, ``Effective approaches to attention-based
  neural machine translation,'' in \emph{EMNLP}, 2015, pp. 1412--1421.

\bibitem{williams2003modeling}
B.~M. Williams and L.~A. Hoel, ``Modeling and forecasting vehicular traffic
  flow as a seasonal arima process: Theoretical basis and empirical results,''
  \emph{Journal of Transportation Engineering}, vol. 129, no.~6, pp. 664--672,
  2003.

\bibitem{kingma2014adam}
D.~P. Kingma and J.~Ba, ``Adam: A method for stochastic optimization,''
  \emph{arXiv preprint arXiv:1412.6980}, 2014.

\bibitem{DBLP:conf/icassp/YuZ04}
G.~Yu and C.~Zhang, ``Switching {ARIMA} model based forecasting for traffic
  flow,'' in \emph{ICASSP}, 2004, pp. 429--432.

\bibitem{DBLP:journals/jits/ChandraA09}
S.~R. Chandra and H.~Al{-}Deek, ``Predictions of freeway traffic speeds and
  volumes using vector autoregressive models,'' \emph{IEEE Transactions on
  Intelligent Transportation Systems}, vol.~13, no.~2, pp. 53--72, 2009.

\bibitem{DBLP:journals/cacie/XieZY07}
Y.~Xie, Y.~Zhang, and Z.~Ye, ``Short-term traffic volume forecasting using
  kalman filter with discrete wavelet decomposition,'' \emph{Comput. Aided Civ.
  Infrastructure Eng.}, vol.~22, no.~5, pp. 326--334, 2007.

\bibitem{DBLP:journals/ijon/FuMLL16}
H.~Fu, H.~Ma, Y.~Liu, and D.~Lu, ``A vehicle classification system based on
  hierarchical multi-svms in crowded traffic scenes,'' \emph{Neurocomputing},
  vol. 211, pp. 182--190, 2016.

\bibitem{DBLP:conf/icdm/MayHKSS08}
M.~May, D.~Hecker, C.~K{\"{o}}rner, S.~Scheider, and D.~Schulz, ``A
  vector-geometry based spatial knn-algorithm for traffic frequency
  predictions,'' in \emph{ICDM Workshops}, 2008, pp. 442--447.

\bibitem{DBLP:conf/aaai/ZhangZQ17}
J.~Zhang, Y.~Zheng, and D.~Qi, ``Deep spatio-temporal residual networks for
  citywide crowd flows prediction,'' in \emph{Proceedings of the Thirty-First
  {AAAI} Conference on Artificial Intelligence,{AAAI}}, 2017, pp. 1655--1661.

\bibitem{DBLP:conf/sdm/YuLSDL17}
R.~Yu, Y.~Li, C.~Shahabi, U.~Demiryurek, and Y.~Liu, ``Deep learning: {A}
  generic approach for extreme condition traffic forecasting,'' in
  \emph{Proceedings of the 2017 {SIAM} International Conference on Data
  Mining}, 2017, pp. 777--785.

\bibitem{cuitraffic19}
Z.~Cui, K.~Henrickson, R.~Ke, and Y.~Wang, ``Traffic graph convolutional
  recurrent neural network: A deep learning framework for network-scale traffic
  learning and forecasting,'' \emph{IEEE Transactions on Intelligent
  Transportation Systems}, 2019.

\bibitem{chen2019multi}
W.~Chen, L.~Chen, Y.~Xie, W.~Cao, Y.~Gao, and X.~Feng, ``Multi-range attentive
  bicomponent graph convolutional network for traffic forecasting,''
  \emph{AAAI}, 2020.

\bibitem{DBLP:journals/corr/BahdanauCB14}
D.~Bahdanau, K.~Cho, and Y.~Bengio, ``Neural machine translation by jointly
  learning to align and translate,'' in \emph{ICLR}, 2015.

\bibitem{DBLP:conf/aaai/ShenZLJPZ18}
T.~Shen, T.~Zhou, G.~Long, J.~Jiang, S.~Pan, and C.~Zhang, ``Disan: Directional
  self-attention network for rnn/cnn-free language understanding,'' in
  \emph{Proceedings of the Thirty-Second {AAAI} Conference on Artificial
  Intelligence,{AAAI}}, 2018, pp. 5446--5455.

\bibitem{DBLP:conf/nips/MnihHGK14}
V.~Mnih, N.~Heess, A.~Graves, and K.~Kavukcuoglu, ``Recurrent models of visual
  attention,'' in \emph{Annual Conference on Neural Information Processing
  Systems, {NIPS}}, 2014, pp. 2204--2212.

\bibitem{DBLP:conf/coling/FengHYZ16}
J.~Feng, M.~Huang, Y.~Yang, and X.~Zhu, ``{GAKE:} graph aware knowledge
  embedding,'' in \emph{26th International Conference on Computational
  Linguistics, {COLING}}, 2016, pp. 641--651.

\bibitem{DBLP:conf/ijcai/BaiYK0S19}
L.~Bai, L.~Yao, S.~S. Kanhere, X.~Wang, and Q.~Z. Sheng, ``Stg2seq:
  Spatial-temporal graph to sequence model for multi-step passenger demand
  forecasting,'' in \emph{IJCAI}, 2019, pp. 1981--1987.

\end{thebibliography}

\begin{IEEEbiography}[{\includegraphics[width=1in,height=1.25in,clip,keepaspectratio]{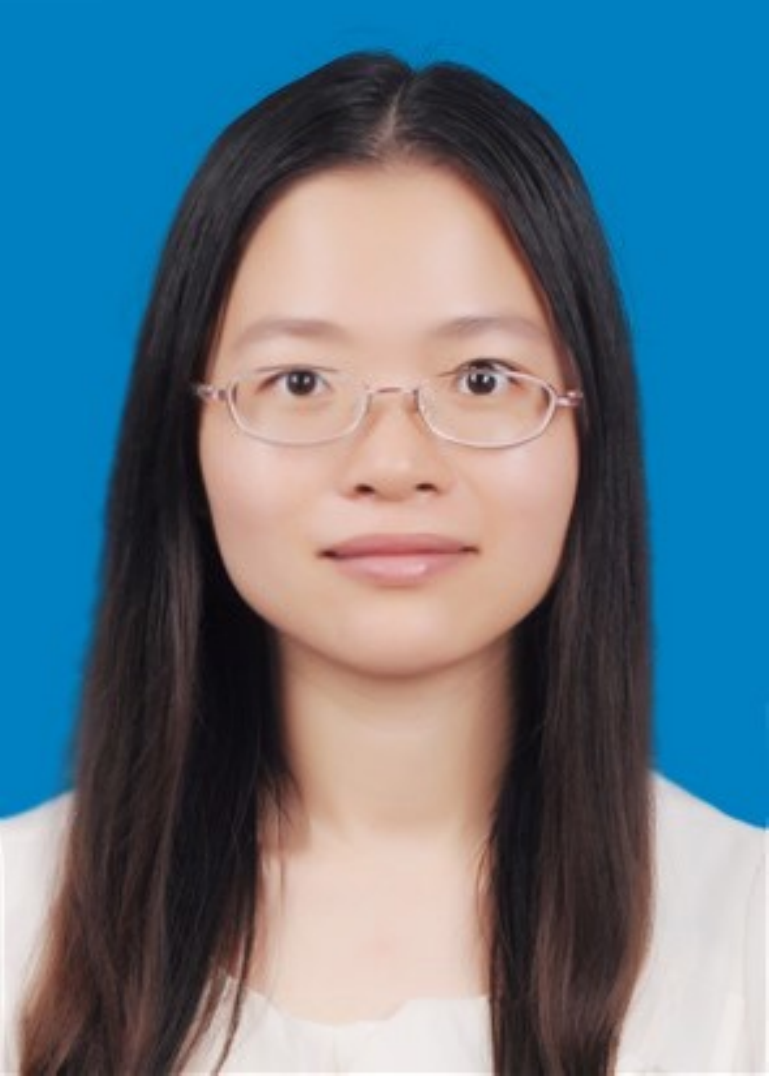}}]{Jiexia Ye}
received the Bachelor's degree in Economics from Sun Yat-sen University in 2012 and M.S. degree in Engineering in Shenzhen Institutes of Advanced Technology, Chinese Academy of Sciences in 2021. Her research interests include graph neural networks / graph embedding in traffic and finance domain.
\end{IEEEbiography}
\begin{IEEEbiography}[{\includegraphics[width=1in,height=1.25in,clip,keepaspectratio]{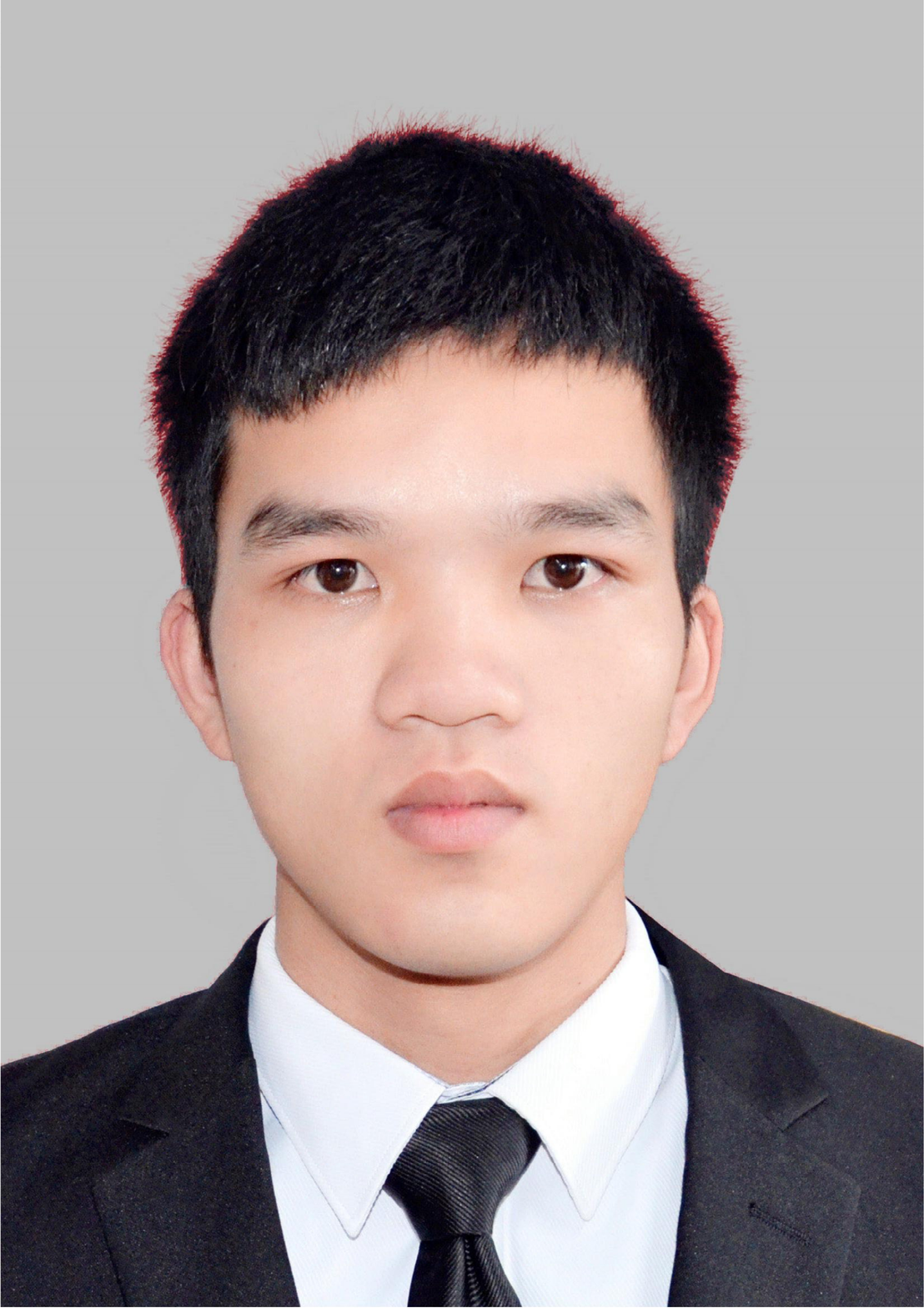}}]{FuRong Zheng}
 received his Bachelor's degree in Automation from Wuyi University in 2018. He is currently working toward M.S. degree  in Computer Science in Shenzhen Institutes of Advanced Technology, Chinese Academy of Sciences. His principal research interest covers the data science and machine learning, in particular, the following areas: traffic analysis; crowd flow prediction and sequential pattern mining.
\end{IEEEbiography}
\begin{IEEEbiography}[{\includegraphics[width=1in,height=1.25in,clip,keepaspectratio]{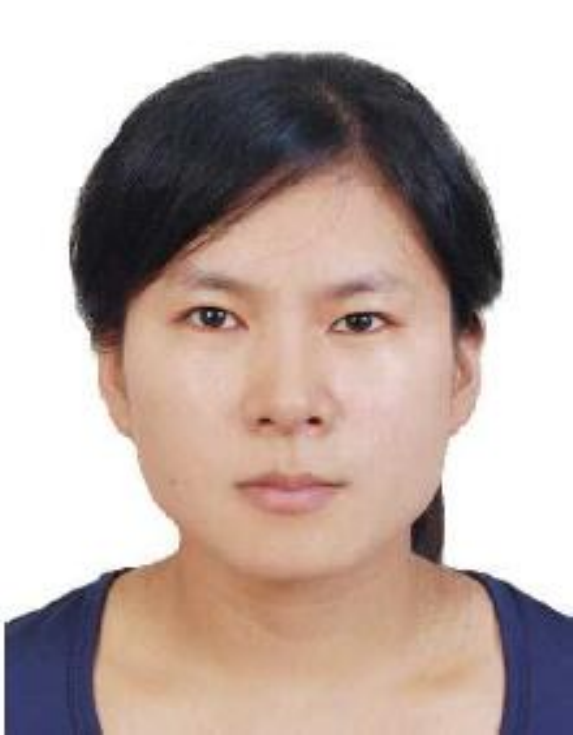}}]{Juanjuan Zhao} received her Ph.D degree from Shenzhen College of Advanced Technology, University of Chinese Academy of Sciences in 2017, and received the M.S. degree from the Department of Computer Science, Wuhan University of Technology in 2009. She is an Assistant Professor at Shenzhen Institutes of Advanced Technology, Chinese Academy of Sciences. Her research topics include data-driven urban systems, mobile data collection, cross-domain data fusion, heterogeneous model integration.
\end{IEEEbiography}
\begin{IEEEbiography}[{\includegraphics[width=1in,height=1.25in,clip,keepaspectratio]{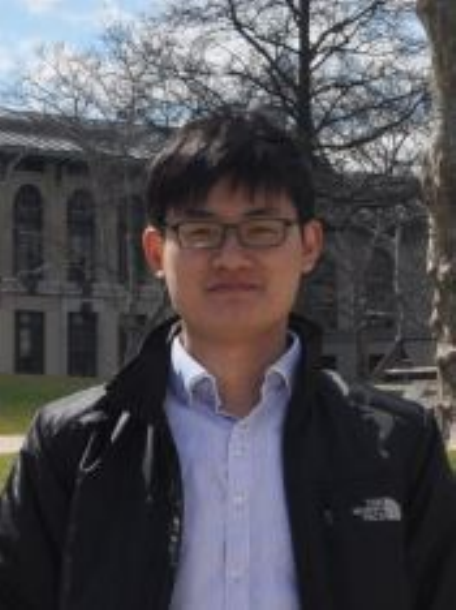}}]{Kejiang Ye} received his BSc and Ph.D degree in Computer Science from Zhejiang University in 2008 and 2013 respectively. He was also a joint Ph.D student at The University of Sydney from 2012 to 2013. After graduation, he worked as Post-Doc Researcher at Carnegie Mellon University from 2014 to 2015 and Wayne State University from 2015 to 2016. He is currently a Professor at Shenzhen Institutes of Advanced Technology, Chinese Academy of Science. His research interests include cloud computing, big data and network systems.
\end{IEEEbiography}
\begin{IEEEbiography}[{\includegraphics[width=1in,height=1.25in,clip,keepaspectratio]{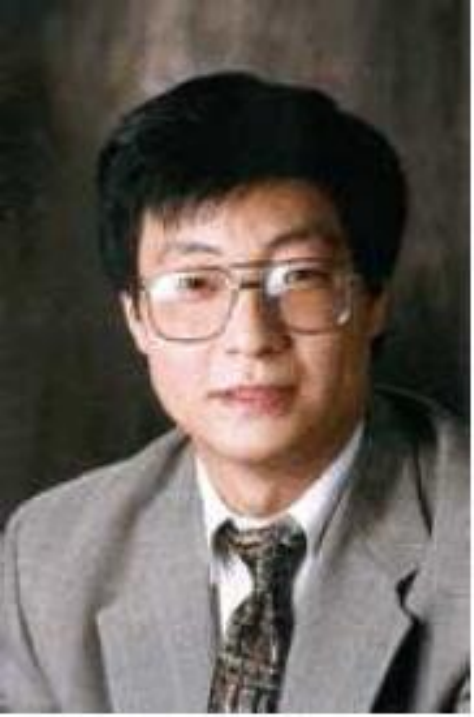}}]{Chengzhong Xu} received his Ph.D degree from the University of Hong Kong, China in 1993. He is the Dean of the Faculty of State Key Lab of IOTSC, Department of Computer Science, University of Macau, Macao SAR, China and a Chair Professor of Computer Science of UM. He was a Chief Scientist of Shenzhen Institutes of Advanced Technology (SIAT) of Chinese Academy of Sciences and the Director of Institute of Advanced Computing and Digital Engineering of SIAT.  He was also in the faculty of Wayne State University, USA for 18 years. Dr. Xu's research interest is mainly in the areas of parallel and distributed systems, cloud and edge computing, and data-driven intelligence. He has published over 300 peer-reviewed papers on these topics with over 10K citations. Dr. Xu served in the editorial boards of leading journals, including IEEE Transactions on Computers, IEEE Transactions on Cloud Computing, IEEE Transactions on Parallel and Distributed Systems and Journal of Parallel and Distributed Computing. He is the Associate Editor-in-Chief of ZTE Communication. He is IEEE Fellow and the Chair of IEEE Technical Committee of Distributed Processing.
\end{IEEEbiography}

\end{document}